\documentclass[letterpaper]{article} 
\usepackage[draft]{aaai2026}  
\usepackage{times}  
\usepackage{helvet}  
\usepackage{courier}  
\usepackage[hyphens]{url}  
\usepackage{graphicx} 
\usepackage{dsfont}
\usepackage{subcaption}  
\usepackage{caption}
\usepackage{comment} 
\usepackage{natbib}  
\usepackage{caption} 
\usepackage{algorithm}
\usepackage[noend]{algpseudocode}
\newcommand{\xunobs}{\mathbf{x}_{\text{unobs}}}
\newcommand{\xobs}{\mathbf{x}_{\text{obs}}}

\usepackage{newfloat}
\usepackage{listings}
\usepackage{amsmath}
\usepackage{amssymb}
\usepackage{pifont}
\usepackage{pgfplots}
\pgfplotsset{compat=1.18} 

\newcommand{\cmark}{~\ding{51}~}%
\newcommand{\xmark}{~\ding{55}~}%

\newtheorem{definition}{Definition}
\DeclareCaptionStyle{ruled}{labelfont=normalfont,labelsep=colon,strut=off} 
\floatstyle{ruled}
\newfloat{listing}{tb}{lst}{}
\floatname{listing}{Listing}
\title{Co-Exploration and Co-Exploitation via Shared Structure in Multi-Task Bandits}
\author{
Sumantrak Mukherjee\textsuperscript{\rm 1}
, Serafima Lebedeva\textsuperscript{\rm 1,2}, Valentin Margraf\textsuperscript{\rm 3,4}, Jonas Hanselle\textsuperscript{\rm 3,4}, Kanta Yamaoka\textsuperscript{\rm 1,2}, Viktor Bengs\textsuperscript{\rm 1}, Stefan Konigorski\textsuperscript{\rm 5,6}, Eyke Hüllermeier\textsuperscript{\rm 1,3,4}, Sebastian Josef Vollmer\textsuperscript{\rm 1,2} 
}
\affiliations{
    \textsuperscript{\rm 1} Department of Data Science and its Applications, German Research Centre for Artificial Intelligence (DFKI), Germany\\
    \textsuperscript{\rm 2}
    Department of Computer Science, University of Kaiserslautern–Landau (RPTU), Germany\\
    \textsuperscript{\rm 3} Institute of Informatics, University of Munich (LMU), Germany\\
    \textsuperscript{\rm 4}
    Munich Center for Machine Learning (MCML), Germany\\
    \textsuperscript{\rm 5} Digital Health – Machine Learning Research Group, Hasso Plattner Institute for Digital Engineering, Potsdam, Germany\\
    \textsuperscript{\rm 6} 
    Hasso Plattner Institute for Digital Health at Mount Sinai, Icahn School of Medicine at Mount Sinai, New York, NY, USA
}

\usepackage{bibentry}

\begin{document}

\maketitle

\begin{abstract}
We propose a novel Bayesian framework for efficient exploration in contextual multi-task multi-armed bandit settings, where the context is only observed partially and dependencies between reward distributions are induced by latent context variables. In order to exploit these structural dependencies, our approach integrates observations across all tasks and learns a global joint distribution, while still allowing personalised inference for new tasks. In this regard, we identify two key sources of epistemic uncertainty, namely structural uncertainty in the latent reward dependencies across arms and tasks, and user-specific uncertainty due to incomplete context and limited interaction history. To put our method into practice, we represent the joint distribution over tasks and rewards using a particle-based approximation of a log-density Gaussian process. This representation enables flexible, data-driven discovery of both inter-arm and inter-task dependencies without prior assumptions on the latent variables. Empirically, we demonstrate that our method outperforms baselines such as hierarchical model bandits, especially in settings with model misspecification or complex latent heterogeneity.
\end{abstract}


\section{Introduction}

Multi-armed bandits (MABs) are a foundational framework for sequential decision-making~\citep{thompson1933_likelihood}, widely used in applications such as personalised recommendation~\citep{zhou2017large} or clinical decision support~\citep{durand2018contextual,shrestha2021bayesian,aziz2021multi}. In these settings, each user represents a new instance of a bandit problem, where the goal is to quickly identify the best arm (e.g., treatment or recommendation) while balancing exploration and exploitation. This becomes especially challenging in heterogeneous populations, where users differ in their optimal choices.

In this work, we consider the multi-task contextual bandit problem~\citep{deshmukh2017multi}
with a partially observable task-context. Often, contextual features such as demographics or test results (\(\xobs\)) are available and can guide personalisation by enabling data sharing across users. However, many important factors, including genetic traits, lifestyle, and long-term adherence, are not directly observed (\(\xunobs\)). This creates a setting with partially informative context, complicating the decision of whether to trust past experience or gather more user-specific information.

This trade-off is particularly critical in clinical applications, where users are patients and the arms correspond to medical treatments. Exploring suboptimal treatments can result in harm or reduced trust, leading to patient dropout or non-adherence. Conversely, over-reliance on incomplete context can cause long-term regret, especially when similar observed features mask underlying heterogeneity.
\begin{figure}[t]
    \centering
    \includegraphics[width=1\linewidth]{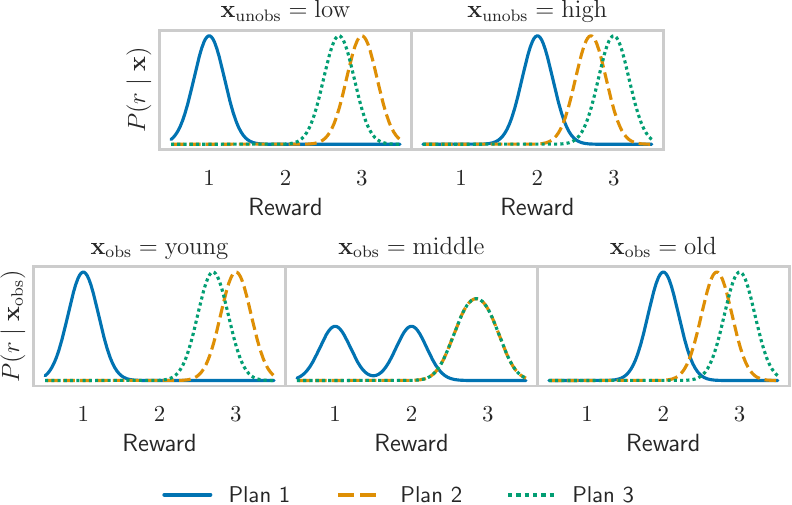}
    \caption{Exemplary reward distributions conditioned on the unobservable feature \emph{metabolic type} (first row) and the observable feature \emph{age group} (second row). }
    \label{fig:contextbasedexploration}
\end{figure}

As a motivating example, consider a system that recommends one of three dietary plans (Figure~\ref{fig:contextbasedexploration}). Each user is defined by an observed feature (age group: young, middle-aged, or old) and an unobserved feature (metabolic type: low or high). We assume that younger users are more likely to have high metabolism, while older users are more likely to have low metabolism. Thus, the observable age group is informative of the unobserved metabolic type. However, for middle-aged users, metabolism may depend on lifestyle factors, and we assume both metabolic types are equally likely for this age group. In Figure~\ref{fig:contextbasedexploration}, we see that Plan 2 is best for users with low metabolism, whereas Plan 3 is preferable for those with high metabolism. Since only the age group is observed, the choice is clear for young and old users, but remains ambiguous for the middle-aged. However, for this group, Plan 1, while always suboptimal, provides valuable information: rewards close to one suggest low metabolism, while rewards near two suggest high metabolism. Thus, assigning Plan 1 to middle-aged users can help identify their metabolic type and, in turn, guide better recommendations between Plans 2 and 3. This example illustrates the importance of context-sensitive exploration strategies that adapt to latent uncertainty. Our method addresses this by modelling the joint reward distribution over users and actions, leveraging shared latent structure to make informed decisions even when some user characteristics are unobserved.

\paragraph{Related work.}
Our contribution intersects several areas of contextual bandit research, including multi-task bandits~\citep{wang2021multitask, deshmukh2017multi}, hierarchical models~\citep{hong2022hierbayesbandits}, meta-learning in bandits~\citep{bastani2022meta, peleg2022metalearning, basu21noregrets,cella2020meta, ortega2019meta}, and mixed-effects modelling for information sharing~\citep{aouali2023mets, huch2024rome}. It also relates to work addressing partial observability~\citep{wang2016learning_missing} and cold-start scenarios~\citep{Bharadhwaj2019metalearningforcoldstart, silva2023user}. These works address the presence of latent variables and information sharing across bandit instances in the presence of uncertainty or partial observability. However, most approaches make strong assumptions about the structure of missing variables, the form of context–reward relationships, or how information is shared across tasks. Assumptions such as rigid clustering, sequential task completion, or static recruitment, limit their flexibility in dynamic, heterogeneous environments. In contrast, our method relaxes these assumptions by modelling a nonparametric distribution over latent reward functions, conditioned on partially observed context. This enables dynamic task recruitment, concurrent interactions, and robust adaptation without relying on predefined structure. 
A comprehensive review of related literature is provided in the Appendix.

We summarize the key distinctions between our model and the most relevant prior work consisting of hLinUCB~\citep{wang2016learning_missing}, KMTL-UCB~\citep{deshmukh2017multi}, GradBand~\citep{kveton2021meta}, RobustAgg~\citep{wang2021multitask}, MTTS~\citep{wan21mtts}, HierTS~\citep{hong2022hierbayesbandits}, and RoME~\citep{huch2024rome} in Table~\ref{tab:novelty_comparison}, which compares six criteria: (i) use of multi-task learning, (ii) support for partial context, (iii) nonparametric treatment of reward structure, (iv) explicit modelling of uncertainty sources, (v) dynamic task recruitment, and (vi) concurrent interaction across tasks.

Our contributions can be summarised as follows:

\begin{itemize}
    \item We propose a unified framework for contextual multi-task, multi-armed bandits that enables more efficient exploration through learning latent information-sharing structures between users and actions, which we call CoCo Bandits.

    \item We distinguish and model three sources of uncertainty: uncertainty over task and action structure, user-specific uncertainty, and uncertainty due to reward noise.
    \item We introduce two novel arm pulling strategies that extend existing approaches by 1) exploiting the learned structure of information sharing between arms, while accounting for uncertainty in that structure, and 2) trading off between learning a global structure that benefits all users and minimising individual user regret.

    \item We demonstrate the effectiveness of our approach on synthetic benchmarks, showing improved adaptability and regret performance through context-sensitive exploration and collaborative structure learning.
\end{itemize}

\begin{table}[t]
\centering
\begin{tabular}{|l|c|c|c|c|c|c|}
\hline
& MT & PC & 
NPLRS
& UD
& DR
& CI
\\
\hline

hLinUCB
& \xmark & \cmark & \xmark  & \xmark & \xmark & \xmark \\

KMTL-UCB  
& \cmark & \xmark & n/a & \xmark & \xmark  & \xmark \\

GradBand 
& \xmark & \xmark & n/a & \xmark& \xmark   & \xmark \\

RobustAgg
& \cmark & n/a 
& n/a  & \xmark & \xmark & \cmark \\
MTTS
& \cmark & \xmark & \xmark  & \xmark& \xmark  & \cmark \\
HierTS 
& \cmark & n/a & \xmark & \xmark & \cmark & \cmark \\
RoME
& \cmark & \xmark & n/a& \xmark& \cmark & \cmark \\

\textbf{CoCo Bandits} & \cmark & \cmark & \cmark & \cmark &\cmark & \cmark \\
\hline
\end{tabular}
\caption{Comparison of our considered setting to related work regarding different criteria: Multi-task setting (MT), partial context (PC), nonparametric latent reward structures (NPLRS), explicit uncertainty distinction (UD), dynamic recruitment of users (DR), where new users can join the existing decision-making process flexibly, and concurrent interaction (CI) among users and algorithms. We use \cmark to indicate that a criterion is met, \xmark if it is unmet, and n/a when the criterion is not applicable.}
\label{tab:novelty_comparison}
\end{table}

\paragraph{Notation.}
We denote probability distributions over random variables \( x \) as \( P(x) \), and their evaluation at specific values as \( P(x = x_0) \) or \( P(x_0) \). Random variables appear unindexed (e.g., \( x, \mu \)), while subscripts (e.g., \( x_i, \mu_j \)) indicate user- or time-specific realizations. Bold symbols (e.g., \( \boldsymbol{x} \)) represent (random) vectors.

Joint and conditional distributions are written as \( P(x, y) \), \( P(x \mid y=y_i) \). When a distribution depends on a latent function or parameter \( f \), we write \( P^{(f)}(x) \) or \( P^{(f)}(x \mid y) \). Distributions over such functions or distributions are denoted by \( Q \), e.g., \( Q(f) \) or \( Q(P) \), depending on the context.

\section{Problem Setting}

We consider a \emph{multi-task contextual bandit} setting, 
~\citep{langford_neurips07}
where tasks, or more specifically users, arrive over time and a shared algorithm selects actions for them. 

Users are indexed by \( i = 1, 2, \dots \) and are recruited in batches of size $m$. At each global time step \( t = 1, 2, \dots \), the algorithm serves exactly one user from the current active set \( I_t \) (which we define below), in a round robin fashion, and the global index \( t \) increases by one for each individual user interaction. 
Every user has a predetermined fixed number of interaction steps denoted by $T_\text{user}$.
At any global time step $t$, the total number of time steps user $i$ interacted with the algorithm is given by $t_i$ and the active set of users $I_t = \{ i \colon t_i < T_\text{user}\}$ consists of those users who have not yet completed their allocated number of interaction steps with the algorithm.
Once every active user has been served (completing a full round), a new batch is recruited from the true population distribution \( P^*(\boldsymbol{x}) \), and added to \( I_t \). Users who have completed their allocated interaction horizon \( T_{\text{user}} \) are removed from \( I_t \), and this process is repeated. Note that while the global time step increases with every user interaction, new users are only recruited once every \( m_t \) global steps, i.e., after all \( m_t = |I_t| \) currently active users have each been served once. The full recruitment and interaction scheme is detailed in Algorithm~\ref{alg:coco}.
We adopt this regular scheme for simplicity, though our framework supports more general recruitment schemes and is robust to the specific choice.

Each user \( i \) is associated with a fixed context vector
\[
\boldsymbol{x}_i = (\boldsymbol{x}_{i,\text{obs}},\boldsymbol{x}_{i,\text{unobs}}) \in \mathbb{R}^d,
\]
sampled i.i.d.\ from \( P^*(\boldsymbol{x}) \). The observed component \( \boldsymbol{x}_{i,\text{obs}}\) is available to the algorithm, while the unobserved component \( \boldsymbol{x}_{i,\text{unobs}}\) captures latent heterogeneity across users, similar to \cite{wang2016learning_missing}. 
Importantly, this context vector \( \boldsymbol{x}_i \) remains fixed for all interaction rounds of user \( i \), unlike standard contextual bandit settings 

\cite{li2010contextual, huch2024rome} where context may vary per decision.
For the user \( i(t) \in I_t \) served at time \( t \), the algorithm selects an arm
\(
a_t \in [K] = \{1, 2, \dots, K\},
\)
and observes a noisy reward
\[
r_t = r_{i(t), t} = \mu_{i(t), a_t} + \epsilon_t,
\]
where \( \mu_{i,a} = g_a(\boldsymbol{x}_i) \) is the latent reward mean for arm \( a \) and \( \epsilon_t \sim \mathcal{N}(0, \sigma^2) \) represents observation noise. The true mean rewards for all arms for user $i$ is given by $\boldsymbol{\mu}_i=(\mu_{i,a})_{a\in[K]}$.

The global history of all interactions up to time \( t \) is
\[
\mathcal{H}_t = \{ (r_s, a_s, i(s)) \}_{s=1}^t,
\]
while the history of a particular user \( j \) at time \( t \) is
\[
\mathcal{H}_{j,t} = \{ (r_s, a_s) : i(s) = j,\, s \le t \}.
\]

In this multi-task setting, our goal is to pull arms for different users, such that we minimise expected cumulative regret. We investigate \textit{Bayes regret}, as typically done \citep{wan21mtts}, and also use a form of \textit{multi-task regret}~\citep{wan21mtts}, defined in Section \ref{sec:regret}.

\subsection{Sources of Uncertainty}
\label{sec:uncertainty}
We distinguish the following sources of uncertainty~\citep{hullermeier2021aleatoric} in the model, each arising from different factors and comprising both aleatoric and epistemic uncertainty about the true reward function:

\begin{enumerate}
    \item \textbf{Aleatoric uncertainty:}  
    Even if the true mean vector \( \boldsymbol{\mu}_i = g(\boldsymbol{x}_i)\) were known, observed rewards $r \sim \mathcal{N}(\mu_{i,a}, \sigma)$ remain noisy. This irreducible uncertainty arises from randomness in the reward mechanism.

    \item \textbf{Individual-level epistemic uncertainty:}  
    As the context $\boldsymbol{x}_i$ is only partly known, there is uncertainty about the true mean vector $\boldsymbol{\mu}_i$. This uncertainty is captured by the conditional distribution $P(\boldsymbol{\mu}\mid \boldsymbol{x}_{i,\mathrm{obs}})$, which marginalizes over the unobserved part $\boldsymbol{x}_{i,\mathrm{unobs}}$. 
    The uncertainty over $\boldsymbol{\mu}_i$ is reduced by conditioning the prior $P(\boldsymbol{\mu}\mid \boldsymbol{x}_{i,\mathrm{obs}})$ on user $i$'s history at time $t$ as $P(\boldsymbol{\mu}\mid \boldsymbol{x}_{i,\mathrm{obs}},\mathcal{H}_{i,t})$.
    The distribution  \(P(\boldsymbol{\mu}\mid  \boldsymbol{x}_{i,\mathrm{obs}} ,\mathcal{H}_{i,t})
    \)
    incorporates information about this history and reduces uncertainty through repeated interaction with the user.

    \item \textbf{Population-level epistemic uncertainty:}
    According to what we just said, learning about $\boldsymbol{\mu}_i$ could be done by proceeding from $P(\boldsymbol{\mu}\mid \boldsymbol{x}_{i,\mathrm{obs}})$ or, more generally, the joint distribution $P(\boldsymbol{\mu}, \boldsymbol{x}_{\mathrm{obs}})$ as a prior and turning it into a posterior by conditioning on $\mathcal{H}_{i,t}$. However, the true distribution $P$ is actually not known. Therefore, we model uncertainty about $P$ in terms of another (meta) distribution $Q$, i.e., a \emph{distribution over distributions}: For each distribution $P(\cdot , \cdot)$, the probability (density) $Q(P)$ expresses our belief that $P$ corresponds to the true distribution on (mean vector, observed context) pairs. This population-level uncertainty captures both functional ($g$) and contextual ($\boldsymbol{x}$) uncertainty and is reduced with an increasing number of users.
\end{enumerate}

\subsection{Reward Structure and Information Sharing}

We define \emph{structure} as statistical dependencies between random variables (e.g., arms or users) such that information about one variable reduces uncertainty about another. In our framework, two main types of structure guide information sharing: shared latent structure across arms for a single user, and the meta distribution over latent structure across users.

\begin{definition}[Shared Latent Structure]
The shared latent structure between arms, given context $\boldsymbol{x}_{i,\text{obs}}$ is defined as the joint distribution over rewards: $P(\boldsymbol{\mu}\mid\boldsymbol{x}_{i,\text{obs}})$.

\end{definition}
For a user $i$ with observed context $\boldsymbol{x}_{i,\mathrm{obs}}$, history $\mathcal{H}_{i,t}$ until time $t$, the joint reward distribution is
$P(\boldsymbol{\mu} \mid \boldsymbol{x}_{i,\mathrm{obs}},\mathcal{H}_{i,t})$.

This distribution captures all statistical dependencies between the arms and observed user components and implicitly reflects both the uncertainty over unobserved components of the context and any latent dependencies in the reward-generating mechanism.
\noindent
For a given joint distribution, $P(\boldsymbol{\mu},\boldsymbol{x}_\text{obs})$, if we observe the reward of arm \( a \), the predictive distribution of the reward for another arm \( b \neq a \) is given by:
\[
    P(\mu_b \mid \mu_a, \boldsymbol{x}_{i,\mathrm{obs}},\mathcal{H}_{i,t}) 
    = \tfrac{P(\mu_a, \mu_b \mid \boldsymbol{x}_{i,\mathrm{obs}},\mathcal{H}_{i,t})}
           {P(\mu_a \mid \boldsymbol{x}_{i,\mathrm{obs}},\mathcal{H}_{i,t})}.
\]
These dependencies formalise how the reward of one arm informs predictions for other arms.

\begin{definition}[Meta-Distribution over Latent Structure]
The meta-distribution $Q$ is defined as a distribution over shared latent structures $P = P(\boldsymbol{\mu} , \boldsymbol{x}_{\text{obs}})$, where $Q(P)$ denotes the probability (density) of the distribution $P$.

\end{definition}

This meta-distribution expresses uncertainty over the underlying joint distribution relating rewards and contexts across users. Rather than assuming a fixed prior or parametric form for how rewards relate to context, we model a distribution over possible joint distributions, enabling structure discovery and adaptation in heterogeneous populations. As user data accumulates, our estimate of \(P(\boldsymbol{\mu} ,\boldsymbol{x}_{\mathrm{obs}})\) becomes more precise, reducing population-level uncertainty.

\subsection{Regret and the Value of Shared Structure}
\label{sec:regret}

We evaluate the performance of our algorithm by measuring its cumulative regret across the users (tasks). In particular, we distinguish between the standard Bayes regret and a refined \emph{multi-task regret} that isolates the value of learning and sharing information across users. In the following, we assume $N$ different users, again indexed by $i$, have interacted with the algorithm over $t$ time steps. 

\paragraph{Bayes regret.}

The Bayes regret of a multi-task bandit algorithm \citep{wan21mtts}
is defined as
\begin{equation}
\text{BR}(t) = \mathbb{E}_{ \boldsymbol{x}, r,\epsilon} \Big[
    \sum\nolimits_{s=0}^t
    \big( \max\nolimits_{a \in [K]} r_{i(s),a} - r_{s} \big)
\Big] \, .
\label{eq:bayes_regret}
\end{equation}

The expectation is taken over the random draw of the reward model \( P \sim Q(P) \) at any given time step, the user contexts \( \boldsymbol{x}_i = \{x_i\}_{i=1}^N \), the reward noise, the interaction order, and the internal randomness of the algorithm. Without information sharing across users, this reduces to the standard single-task Bayes regret over \( N \) independent problems.

\paragraph{Multi-task regret.}

To measure the value of learning shared structure, we define the \emph{multi-task regret} \citep{wan21mtts} as the gap between the performance of the algorithm and a policy with access to the oracle joint distribution over the population that has access to the true conditional reward distribution \(P^*(\boldsymbol{\mu}\mid\boldsymbol{x}_\text{obs})\) for each user. This oracle uses Thompson sampling with the true prior and selects actions accordingly:
\begin{equation}
a_{t}^{\text{oracle}} \sim \text{TS}\left( P^*(\boldsymbol{\mu}\mid\boldsymbol{x}_{i,\text{obs}},\mathcal{H}_{i(t),\text{oracle}}) \right) \, ,
\label{eq:oracle_ts}
\end{equation}
where $\mathcal{H}_{i(t),\text{oracle}}$ is the history collected by a Thompson sampling algorithm with oracle prior for user $i$.
The multi-task regret is defined as
\begin{equation}
\text{MTR}(t) = \mathbb{E}_{ \boldsymbol{x}, r, \epsilon} \Big[
    \sum\nolimits_{s=0}^t
    ( r_{i,a^{\text{oracle}}_{s}} - r_{s} )
\Big] \, .
\end{equation}

This metric captures the performance gap due to not knowing the true task-specific prior in advance. It quantifies the benefit of learning a meta-model that generalises across users. A well-designed multi-task algorithm should minimise this regret by learning the structure \( P(\boldsymbol{\mu} \mid \boldsymbol{x}_\text{obs}) \) and leveraging it for future decisions.

\section{Methods}
Our method builds on a hierarchical Bayesian framework~\citep{gelman1995bayesian} for multi-task contextual bandits. We begin by introducing a flexible, nonparametric model over latent reward distributions that captures shared structure across users and arms. We then describe how personalised posteriors are inferred using Sequential Monte Carlo (SMC) with a Karhunen–Loève (KL) expansion. Next, we outline how user-specific observations are incorporated into these posteriors via hierarchical Bayesian updates. We then develop acquisition strategies that balance individual-level and population-level uncertainty to guide exploration. Finally, we present Algorithm~\ref{alg:coco}, which enables concurrent, adaptive decision-making across users.

\subsection{Modelling of Reward Distributions}
\label{sec:meta-modelling}

To flexibly model the joint distribution over arm means \( \boldsymbol{\mu} \in \mathbb{R}^K \) and observable user context \( \boldsymbol{x}_{\text{obs}} \in \mathbb{R}^{d_{\text{obs}}} \), we adopt a nonparametric Bayesian approach based on Gaussian Processes (GPs). GPs are well-suited for this task, as they can approximate any continuous function under mild regularity conditions, making them universal density approximators when applied in log-space. This allows us to model a wide range of distributions (multimodal, skewed, heavy-tailed, or correlated) without committing to a fixed parametric form \cite{murray2008gaussian}.

In our setting, we place a GP prior over the log-density of the joint distribution \( P(\boldsymbol{\mu}, \boldsymbol{x}_{\text{obs}}) \). That is, we model:
\[
\log P^{(f)}(\boldsymbol{\mu}, \boldsymbol{x}_{\text{obs}}) = f(\boldsymbol{\mu}, \boldsymbol{x}_{\text{obs}}), \qquad f \sim \mathcal{GP}(m(\cdot), k(\cdot, \cdot))
\]
Exponentiating a sample from the GP defines an unnormalised joint density:
\(
\tilde{P}_f(\boldsymbol{\mu}, \boldsymbol{x}_{\text{obs}}) = \exp(f(\boldsymbol{\mu}, \boldsymbol{x}_{\text{obs}}))
\).
To obtain a valid probability distribution, we normalise:
\[
P^{(f)}(\boldsymbol{\mu}, \boldsymbol{x}_{\text{obs}}) = \tfrac{\exp(f(\boldsymbol{\mu}, \boldsymbol{x}_{\text{obs}}))}{\int \exp(f(\boldsymbol{\mu}, \boldsymbol{x}_{\text{obs}})) \, d\boldsymbol{\mu} \, d\boldsymbol{x}_{\text{obs}}}
\]

This defines a distribution over joint distributions—that is, a meta-prior \( Q(f) \) over reward models \( f \), each of which induces a valid joint density \( P^{(f)}(\boldsymbol{\mu}, \boldsymbol{x}_{\text{obs}})\).

This formulation provides expressive modelling capacity and a natural way to quantify uncertainty over both arm rewards and user context distributions. By modelling the joint density directly, we can capture complex dependencies and share statistical strength across users with similar metadata. This is particularly useful in hierarchical multi-task settings, where each user contributes a few observations, and robust structure learning is essential.

\paragraph{Approximate inference with Karhunen–Loève expansion and Sequential Monte Carlo.}

Modelling distributions over log-densities with Gaussian Processes introduces significant computational challenges: GPs have \( \mathcal{O}(N^3) \) complexity in the number of evaluation points, and repeated conditioning or sampling becomes intractable as user interactions grow. Since users in our setting arrive sequentially and interact with the algorithm over multiple rounds, we adopt a scalable inference strategy based on Sequential Monte Carlo (SMC)~\cite{chopin2020introduction}. Specifically, we approximate GP samples using a truncated Karhunen–Loève (KL) expansion~\cite{qi2012eigengp}, which reduces the infinite-dimensional GP to a finite, low-rank basis representation. Further details, including the full KL construction, softmax normalisation, and SMC update algorithm, are provided in the Appendix.
\subsection{Bayesian Inference on Reward Distributions}
\label{sec:BayesUpdates}

We model population-level uncertainty by placing a prior \( Q(f) \) over reward models \( P^{(f)} \), 
where each plausible joint distribution over arm means and context is parametrised by $f$. Specifically, \( P^{(f)}(\boldsymbol{\mu} \mid \boldsymbol{x}_{i,\text{obs}}) \) denotes the distribution over the arm mean vector \( \boldsymbol{\mu} \in \mathbb{R}^K \) conditioned on user context \(\boldsymbol{x}_{i,\text{obs}}  \) and reward model $f$. 

Given the interaction history \( \mathcal{H}_t \) until time step $t$, the posterior over models is given by Bayes' rule:
\begin{equation}
Q(f \mid \mathcal{H}_t) \propto P(\mathcal{H}_t \mid f) \cdot Q(f)
\end{equation}
where \( P(\mathcal{H}_t \mid f) \) is the probability of the data given \( f \).

For a user $i$ interacting with the algorithm at time $t$, we condition draws $f\sim Q(f)$ from the meta distribution over the personal history $\mathcal{H}_{i,t}$, i.e. $P^{(f)}(\boldsymbol{\mu}\mid \boldsymbol{x}_\text{i,obs},\mathcal{H}_{i,t})$.

To update the meta distribution, we calculate the probability of the observed reward for each $f$, as $Q(f\mid r_t,a_t,\mathcal{H}_t)$.

\paragraph{Single observation probability.} The marginal probability of observing reward \( r \) according to the joint distribution $P^{(f)}(\boldsymbol{\mu},\boldsymbol{x}_\text{obs})$ from pulling arm \( a \) is $P(r \mid f, a, \boldsymbol{x}_\text{i,obs})=$ 
\(
 \int \mathcal{N}(r \mid \mu_{a}, \sigma^2)  \int P^{(f)}(\boldsymbol{\mu} \mid \boldsymbol{x}_\text{i,obs}) \, d\boldsymbol{\mu}_{-a}  d\mu_{a} \, .
\label{eq:marginal_likelihood}
\)

\paragraph{User history probability.} The probability of a user history for a given joint distribution $f$ is
\(
P^{(f)}(\mathcal{H}_{i,t} \mid \boldsymbol{x}_\text{i,obs}) = \int_{\mathbb{R}^K} \left[ \prod_{t\in\mathcal{H}_{i,t}} \mathcal{N}(r_t \mid \mu_{a_t}, \sigma^2) \right] P^{(f)}(\boldsymbol{\mu} \mid \boldsymbol{x}_\text{i,obs}) \, d\boldsymbol{\mu} \, .
\)

\paragraph{Dataset probability.} Assuming users are conditionally independent given \( f \), the probability of the dataset is
\[
P(\mathcal{H}_t \mid f) = \prod\nolimits_{i \in \mathcal{I}} P^{(f)}(\mathcal{H}_{i,t} \mid  \boldsymbol{x}_\text{i,obs}) \, .
\]

\paragraph{Posterior update with new observation.} Let $Q_t(f)=Q(f\mid\mathcal{H}_t)$ denote the likelihood of the joint distribution induced by $f$ after having interacted with the algorithm $t$ time steps and collected data $\mathcal{H}_t$. A new interaction \( (a, r) \) for user \( i \), updates the model posterior to $Q_{t+1}(f)$:
\begin{equation*}
Q(f \mid \mathcal{H}_{i,t}, \boldsymbol{x}_\text{i,obs}, a, r) \propto 
P(r \mid f, a, \boldsymbol{x}_\text{i,obs}, \mathcal{H}_{i,t})
\cdot 
Q_t(f)
\end{equation*}
where we can express the latter posterior as 
\begin{equation*}
\int \mathcal{N}(r \mid \mu_a, \sigma^2) \cdot P^{(f)}(\boldsymbol{\mu} \mid \boldsymbol{x}_\text{i,obs}, \mathcal{H}_{i,t}) \, d\boldsymbol{\mu}_{-a}  d\mu_{a}
\end{equation*}

This formulation enables fully Bayesian updates over model space and supports structure-aware reasoning conditioned on both user history and shared population structure.

\subsection{Exploration Under Meta-Level Uncertainty}

\label{sec:acquisition}

In hierarchical multi-task bandits, exploration must adapt to both individual uncertainty and population-level uncertainty over shared structure as described in Section~\ref{sec:uncertainty}.

A fully Bayesian, nonparametric approach enables us to quantify both forms of uncertainty without strong parametric assumptions. This flexibility is crucial in heterogeneous environments: it allows the model to rapidly adapt to new or atypical users without overcommitting to prior structure, while still leveraging learned correlations when they are present. The posterior over joint reward distributions evolves as more users interact, allowing the agent to exploit structure when reliable and fall back to local learning otherwise.

\paragraph{Balancing local and global gains.}

Effective exploration should serve one of two goals: (i) reduce uncertainty about the current user to minimise regret, or (ii) improve the global reward model to benefit future users. Our acquisition strategy explicitly balances both these goals.

\subsection{Acquisition Functions}

Our framework employs two acquisition strategies built on the shared population-level structure. First, Thompson sampling with a nonparametric meta-prior (TS-NP) leverages a flexible, nonparametric prior to model heterogeneous subpopulations, quantifying uncertainty through both user history and global information without overbiasing towards other users.
Second, global information directed sampling (GIDS) targets population-level learning by actively reducing uncertainty in the meta-prior, enabling collaborative exploration that benefits future users.

\paragraph{Thompson sampling.}

We define the optimal arm distribution for user $i$ at time $t$ under a given model \( f \) by conditioning the prior $P^{(f)}$ on user context and history as  $P_{t}^{(i,f)}(\boldsymbol{\mu}) =P^{(f)}(\boldsymbol{\mu} \mid \boldsymbol{x}_{i,\text{obs}},\mathcal{H}_{i,t})$. Then, 
\[
P(a^* = k\mid \boldsymbol{x}_{i,\text{obs}},\mathcal{H}_{i,t},f) = P_{P_{t}^{(i,f)}} \left( \arg\max\nolimits_a \mu_a = k \right) \, .
\]
The marginal best-arm distribution, which accounts for uncertainty over models via the meta-prior \( Q_t(f) \), is given by
\[
P_t^{(i)}(a^* = k ) = \mathbb{E}_{f \sim Q_t(f)} \big[ P^{(f)}(a^* = k\mid \boldsymbol{x}_{i,\text{obs}},\mathcal{H}_{i,t}) \big] \, 
\]
We employ a Thompson Sampling strategy extended to the hierarchical setting. This approach leverages population-level information by utilising the posterior $Q_t(f)$, while also incorporating user-specific information.
Sampling \( f \sim Q_t(f) \), we compute the probability of each arm \( k \) being optimal for user $i$ as $P^{(f)}(a^* = k\mid \boldsymbol{x}_{i,\text{obs}},\mathcal{H}_{i,t})$
and then sample the action as
\(
a_{i,t}=k \sim P^{(f)}(a^* = k \mid \boldsymbol{x}_{i,\text{obs}} , \mathcal{H}_{i,t}).
\) This procedure enables the agent to balance shared knowledge across users (via the meta-distribution) and personalised adaptation (via user-specific posteriors).

\paragraph{Global information directed sampling.}
We propose another acquisition function inspired by Information-Directed Sampling (IDS) \cite{russo2014learning}, aimed at reducing uncertainty over the meta-distribution. Unlike standard IDS, which targets individual-level uncertainty about the optimal arm, our approach prioritises reducing uncertainty over the shared latent structure across tasks.
To reduce uncertainty about the population-level structure, we consider information gain over the meta-distribution \( Q_t(f) \) at time $t$. At global time \( t \), the current meta-distribution \( Q_t(f) \) is updated after observing a reward \( r \) from pulling arm \( a \) for user \( i \) with context \( \boldsymbol{x}_{i,\text{obs}} \) and history \( \mathcal{H}_{i,t} \):
\[
Q_t(f \mid r, a, i) \propto P^{(f)}(r \mid a, \boldsymbol{x}_{i,\text{obs}}, \mathcal{H}_{i,t}) \, Q_t(f).
\]
For fixed \( f \), the expected information gain is the reduction in entropy $H(\cdot)$ of the meta-distribution from pulling arm \( a \) for user $i$:
\[
\text{EIG}_t^{(i,f)}(a) =
\mathop{\mathbb{E}}\limits_{{r_a} \sim P^{(i,f)}_{t}(r_a)}
\big[
  H\!\big( Q_t(f) \big) - H\!\big( Q_t(f \mid r, a, i) \big)
\big] \, ,
\]
where ${r_a}\sim P^{(i,f)}_t(r_a)$ is $r_a\sim\mathcal{N}(\mu_a,\sigma^2)$ for $\boldsymbol{\mu}\sim P^{(i,f)}_t(\boldsymbol{\mu})$.
The expected regret for pulling arm \( a \) for user $i$ under \( f \) is
\[
\text{Regret}_t^{(i,f)}(a) =
\mathop{\mathbb{E}}\nolimits_{\boldsymbol{\mu} \sim P^{(i,f)}_{t}(\boldsymbol{\mu})}
[ \max\nolimits_{k} \mu_k - \mu_a ]  \, .
\]
The GIDS objective for a given \( f \) is defined as
\[
\pi_f^* = 
\arg\min\nolimits_{\pi_f \in \Delta_K}
\tfrac{ \text{Regret}_t^{(i,f)}(\pi_f)^2 }{ \text{EIG}_t^{(i,f)}(\pi_f) } \, ,
\]
where
\(
\text{Regret}_t^{(i,f)}(\pi_f) = \sum_a \pi_{f,a} \, \text{Regret}_t^{(i,f)}(a)
\)
and
\(
\text{EIG}_t^{(i,f)}(\pi_f) = \sum_a \pi_{f,a} \, \text{EIG}_t^{(i,f)}(a).
\)
The final action is chosen by first sampling \( f \sim Q_t(f) \), computing \( \pi_f^* \), and sampling \( a \sim \pi_f^* \). This avoids averaging over \( f \), ensuring that the chosen arm is optimal with respect to a single plausible structural hypothesis drawn from the current posterior.

\paragraph{Concurrent decision making via meta-learning.}
We propose the Co-Exploring and Co-Exploiting Multi-task Bandit (\textsc{CoCo-Bandit}) algorithm, described above. 
Alg.~\ref{alg:coco} implements the ideas developed in Sections~\ref{sec:meta-modelling}--\ref{sec:acquisition}, combining hierarchical modelling, particle-based inference, and structure-aware acquisition functions. At each interaction round, the algorithm updates the posterior over models \( Q_t(f ) \), conditions on the selected user's context and history to form a personalised belief, and selects an action using one of our acquisition functions (e.g., Thompson Sampling or global IDS).

\begin{algorithm}[t]
\caption{\textsc{CoCo-Bandits}} \label{alg:coco}
\textbf{Input:} Meta-prior \( Q(f) \), user horizon \( T_{\text{user}} \),  batch size \( m \), reward noise $\sigma$\\
\textbf{Initialise:} 
    Global history \( \mathcal{H}_0 \gets \emptyset \) \\
    Active users \( I_0 \gets \{1, \dots, m\} \), with observed \( \boldsymbol{x}_{i,\text{obs}} \)  \( \forall i \in I_0 \) \\
    Interaction counters \( c_i \gets 0 \) for all \( i \in I_0 \)
\begin{algorithmic}[1]
\For{global time steps \( t = 1, \dots, T \)}
    \State Select user \( i(t) \in I_{t-1} \) using round-robin order 
    \State Update meta-posterior \( Q(f \mid \mathcal{H}_{t-1}) \)  \Comment{See Sec.~\ref{sec:BayesUpdates}}
    \State Compute posterior with user history \( P^{(f)}(\boldsymbol{\mu} \mid t)\)
    \State Select action \( a_t \in [K] \) and observe reward \( r_t \)
    \State Update global history: \( \mathcal{H}_{t} \gets \mathcal{H}_{t-1} \cup \{ (r_t, a_t, i(t)) \} \)
    \State Increment interaction counter: \( c_{i,t} \gets c_{i,t} + 1 \)

    \If{ \( c_{i,t} = T_{\text{user}} \)} \( I_t \gets I_{t-1} \setminus \{i(t)\} \) \Else \( \, I_t \gets I_{t-1} \) \EndIf

    \If{ \( t \mod m = 0 \)}  \Comment{End of current round}
        \State Recruit new batch \( J \sim P^*(\boldsymbol{x}) \), \( |J| = m \)
        \State Observe \( \boldsymbol{x}_{j,\text{obs}} \) for each \( j \in J \)
        \State \( I_t \gets I_t \cup J ,\{c_j\gets0 \}_{j\in J}, m\gets |I_t|\)
    \EndIf
\EndFor
\end{algorithmic}
\end{algorithm}

\begin{figure*}[t]
    \centering

    \begin{subfigure}[b]{0.49\columnwidth}
        \includegraphics[width=\textwidth]{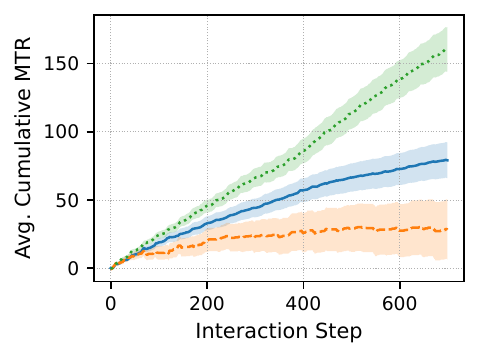}
        \label{fig:6plots_1}
    \end{subfigure}
    \begin{subfigure}[b]{0.49\columnwidth}
        \includegraphics[width=\textwidth]{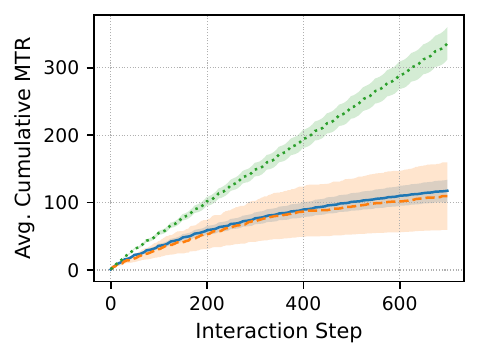}
        \label{fig:6plots_2}
    \end{subfigure}
    \begin{subfigure}[b]{0.49\columnwidth}
        \includegraphics[width=\textwidth]{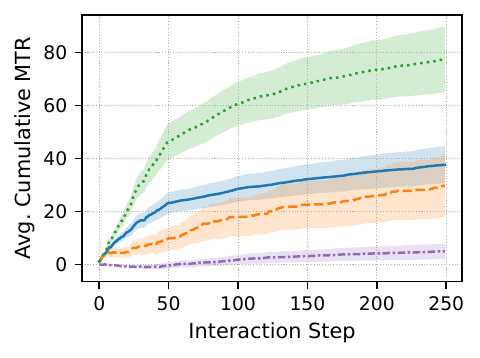}
        \label{fig:6plots_3}
    \end{subfigure}
    \begin{subfigure}[b]{0.49\columnwidth}
        \includegraphics[width=\textwidth]{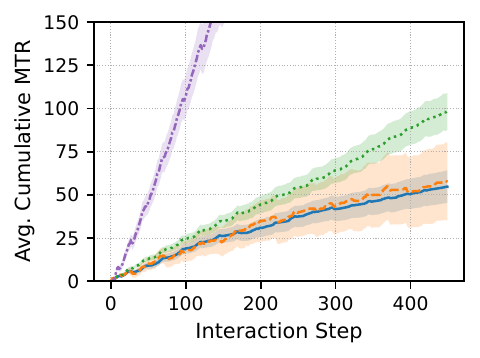}
        \label{fig:6plots_4}
    \end{subfigure}

    \begin{subfigure}[b]{0.49\columnwidth}
        \includegraphics[width=\textwidth]{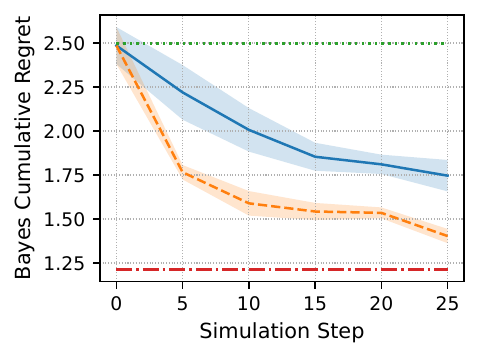}
        \caption{Mixture of Gaussians}
        \label{fig:6plots_5}
    \end{subfigure}
    \begin{subfigure}[b]{0.49\columnwidth}
        \includegraphics[width=\textwidth]{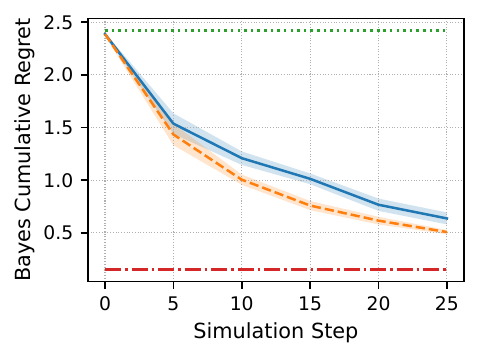}
        \caption{Partially observed context}
        \label{fig:6plots_6}
    \end{subfigure}
    \begin{subfigure}[b]{0.49\columnwidth}
        \includegraphics[width=\textwidth]{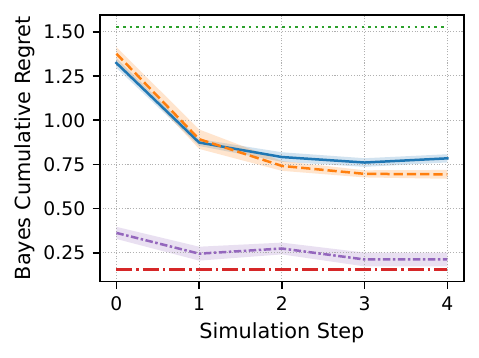}
        \caption{LMM: Gaussian latent}
        \label{fig:6plots_7}
    \end{subfigure}
    \begin{subfigure}[b]{0.49\columnwidth}
        \includegraphics[width=\textwidth]{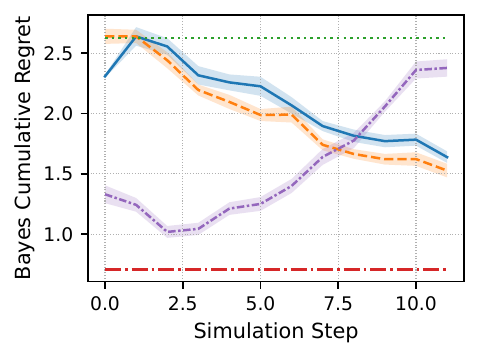}
        \caption{LMM: non-Gaussian latent}
        \label{fig:6plots_8}
    \end{subfigure}
    \vspace{0.0em}
    \begin{tikzpicture}
    \begin{axis}[
        hide axis,
        xmin=0, xmax=1,
        ymin=0, ymax=1,
        legend style={
            draw=none,
            font=\footnotesize,
            legend columns=5, 
            column sep=1.5em,
            at={(0.5,1.0)},
            anchor=south,
        },
    ]
    \addlegendimage{solid, line width=1.2pt}
    \addlegendentry{NPM-TS}
    \addlegendimage{dashed, line width=1.2pt}
    \addlegendentry{GIDS}
    \addlegendimage{dash dot dot, line width=1.2pt}
    \addlegendentry{MTTS}
    \addlegendimage{dotted, line width=1.2pt}
    \addlegendentry{Ind-TS}
    \addlegendimage{dash dot, line width=1.2pt}
    \addlegendentry{OraclePrior-TS}
    \end{axis}
    \end{tikzpicture}

    \vspace{0.1em}

    \caption{
        Comparison of our method with standard baselines and MTTS across four experimental conditions. 
        The \textbf{top row} shows average \emph{multi-task cumulative regret}, capturing population-level performance. 
        The \textbf{bottom row} shows average \emph{Bayesian cumulative regret}, reflecting user-specific regret. 
        The x-axis indicates interaction rounds.
    }
    \label{fig:6plots}
\end{figure*}

\section{Experiments}

We evaluate \textsc{CoCoBandits} (Alg.\ \ref{alg:coco}), equipped with two acquisition strategies, Thompson sampling (\textsc{NPM-TS}) and global information direction sampling  (\textsc{GIDS}) with a nonparametric meta-prior on a series of synthetic bandit environments. Our primary goal is to demonstrate that leveraging a nonparametric meta-prior enables flexible modelling of subpopulation structure, adapts to heterogeneous contexts, and ultimately reduces cumulative regret by sharing information between users.

We compare our approach against \textsc{MTTS} by \citet{wan21mtts}, specifically its Gaussian Linear Mixed Effect Model (LMM) variant, the only suitable benchmark in our setting, as it models context-adaptive priors and maintains a user-specific distribution. Further, we compare against a version of independent Thompson Sampling referred to as \textsc{Ind-TS} with a uniform prior, highlighting the benefits of information sharing. Other baselines, such as linear contextual bandits or hierarchical Bayesian bandits, fail to adaptively share information across users or handle heterogeneous context dependent priors, making them unsuitable for direct comparison. 

The baselines assume different settings regarding parametric assumptions of the reward distribution on context and partial observability of context. To enable a meaningful comparison, we consider three different synthetic environments. We compare against \textsc{Ind-TS} in the first two environments and both baselines in the last two.

In the first environment, we consider a mixture of Gaussians, to illustrate that our nonparametric prior can flexibly adapt to multi-modal structures, which can typically not be modelled by exponential family distributions.

For our second environment, we consider a linear contextual bandit, with a partially observable context, where the context variables are of mixed types (discrete and continuous) and are correlated with the observable context. 
This environment highlights our method’s ability to handle partially specified contexts and a mixed category of latent variables, without explicitly specifying them. 

We consider two cases in our third environment with a linear-mixed model: 1) the latent variable $\theta$ is sampled from a normal distribution, and 2) it is sampled from a mixture of Gaussians.
We use this environment to compare our model against MTTS, as the parametric assumptions for the reward generation are met; however, in the second case we demonstrate that even when the parametric assumptions are fulfilled, misspecifying the distribution of the latent can lead to overfitting to a subgroup and hence yield linear regret. We provide a detailed description of the generation of the synthetic environments and the baselines in the appendix.

\subsection{Results}

In Figure~\ref{fig:6plots}, we report the mean performance averaged over 10 random seeds, along with standard deviations normalised by $\sqrt{10}$. Results are shown separately for each environment: Environment 1 is presented in the first column (Figures~\ref {fig:6plots_5}), Environment 2 in the second column (Figures \ref{fig:6plots_6}), and so on. The first row illustrates the average cumulative multi-task regret at all interaction steps, while the second row presents the average Bayes cumulative regret for a randomly sampled fixed evaluation batch over 10 arm pulling rounds using TS with priors learned by respective algorithms evaluated after every round of active users interactions.

In the mixture of Gaussians setting (Figures~\ref{fig:6plots_5}), our model asymptotically converges to the behaviour of the oracle prior in both cumulative multi-task and Bayes regret. The cumulative regret growth flattens, and the Bayes regret approaches that of the oracle. \textsc{GIDS} outperforms \textsc{NPM-TS}, as expected: by initially exploring informative but suboptimal arms, it learns the latent task structure. Each Gaussian mode corresponds to a user subpopulation with a distinct optimal arm, enabling efficient personalisation through early-round inference for tasks in later time steps.

In the setting with partial context and mixed latent variables (Figures~\ref{fig:6plots_6}), both acquisition functions gradually align with the oracle prior as more users are observed, indicating successful learning of the population structure. However, the performance gap between \textsc{GIDS} and \textsc{NPM-TS} is small in the multi-task regret, which is expected since the structure does not imply clear subgroups with distinct optimal arms, hence exploration is still necessary. The average Bayes cumulative regret shows that leveraging the learned structure improves efficiency, although there is no massive advantage in this case.

In the third experiment (Figures~\ref{fig:6plots_7}),
we follow the setting as proposed by \citet{wan21mtts}
where the latent parameter is sampled from a Gaussian and the environment exactly aligns with MTTS's parametric assumptions.
Following MTTS's settings, we assume that all tasks join together instead of following a dynamic recruitment setting.
MTTS converges fastest, as expected. Still, both \textsc{GIDS} and \textsc{NPM-TS} asymptotically converge to the true population distribution, demonstrating competitive performance even without parametric assumptions. Our methods also benefit from information sharing, and therefore outperform \textsc{Ind-TS}.
In this setting, since we interact with a fixed cohort of users over multiple rounds, the cumulative Bayes regret for the evaluation batch plateaus once the posterior has fully incorporated the available user data.

In the fourth experiment (Figures~\ref{fig:6plots_8}), we again use an LMM but sample $\theta$ from a mixture of Gaussians and consider users joining dynamically in batches. MTTS performs poorly, incurring linear regret and even underperforming \textsc{Ind-TS}, which is expected, as its parametric assumptions are violated. 
We observe that the average cumulative Bayes regret for \textsc{MTTS} initially decreases as the model conditions its uninformative prior on early users, but then constantly increases due to overfitting effects caused by its parametric specification, resulting in a biased and overconfident prior that harms performance on later users.
In contrast, our model converges with more users and approaches oracle-like performance in terms of average cumulative Bayes regret. This setting highlights that incorrect assumptions about the latent distribution can severely degrade performance.

Additional details regarding the experimental setup, hyperparameters chosen, random seeds and additional results are reported in the appendix.

\section{Discussion}

We introduced a general framework for multitask contextual bandits that captures structure across both actions and tasks through a fully Bayesian, nonparametric model. Our approach enables effective information sharing across users, avoiding cold-start issues without overbiasing individual decisions. This makes our approach particularly suitable for real-world scenarios such as personalised medicine or advertising, where the reward structure and population heterogeneity are uncertain.
 
By modelling the joint distribution over arm means and contexts with minimal parametric assumptions, our approach flexibly adapts to a wide range of heterogeneity patterns. When parametric assumptions of the reward structure are violated, our nonparametric prior outperforms models that rely on such assumptions, offering robust uncertainty quantification and adaptation across heterogeneous tasks. While parametric models can be more sample-efficient when correctly specified, incorrect assumptions often lead to worse performance than independent cold-start strategies. 

To make our model computationally tractable, we proposed an efficient approximation using a truncated KL expansion and Sequential Monte Carlo sampling, enabling online updates and action selection.
However, utilising GPs for nonparametrically modelling probability distributions still comes at a significant computational cost, especially when scaling to higher dimensions. 
Computationally efficient approximations, such as Nyström~\cite{wild2021connections}, will be considered in future work.
Besides, there are several other interesting extensions of our work that we plan to explore, such as batched user acquisition functions and combination with parametric assumptions to balance statistical efficiency with modelling flexibility.

\section{Acknowledgment}
We acknowledge funding by the state of Rhineland-Palatinate, Germany.

\bibliography{aaai2026}

\clearpage
\section*{Appendix}
\appendix
\section{Extended Related Work}\label{sec:related_work}

The multi-armed bandit (MAB) framework is often used for sequential decision-making under uncertainty, balancing exploration and exploitation to minimise cumulative regret. The contextual bandit variant enriches this model by incorporating side information, such as user or item features, to guide action selection~\citep{li2010contextual}. These models generalise across contexts and are central to personalised recommendation and adaptive decision systems.

\paragraph{Contextual Bandits and Partial Observability.}
Contextual bandits have been widely studied under both linear~\citep{lu2010contextual} and nonlinear~\citep{zhou2020neural, ban2024neural} reward assumptions. While linear models offer interpretability and regret guarantees, deep learning methods have improved scalability and flexibility in high-dimensional spaces. However, most rely on the realisability assumption, meaning the expected reward is fully determined by the context~\citep{foster2020adapting, krishnamurthy2021adapting}. In contrast, our method allows for randomness in the reward even when conditioned on observed metadata, making it more robust to model mismatch and better suited for heterogeneous environments.

Recent work has explored partial-contextual settings, where only some features are observable (for example, user features without item data). These motivate structured models that infer missing information through shared statistical patterns~\citep{wang2016learning_missing, park2022tspartially, park2024partially}. Our approach shares this philosophy: we model task-specific reward parameters as latent variables conditioned on observable metadata, which enables principled reasoning under uncertainty and incomplete context.

\paragraph{Meta and Multi-Task Bandits.}
Meta-learning in bandits aims to transfer knowledge across tasks, enabling fast adaptation with limited per-task data~\citep{hsu2019empirical, boutilier2020differentiable, kveton2021meta, kveton2021metalearning}. For example, Meta-TS~\citep{kveton2021meta} assumes reward vectors are drawn i.i.d.\ from a shared distribution and uses observed data to estimate a prior. However, most meta-bandit approaches ignore task metadata and assume a metadata-agnostic task distribution. Our method generalises these by leveraging task-specific metadata and supporting arbitrary task arrival orders, increasing flexibility in real-world deployments.

Related are hierarchical bandits, where task-level reward functions are governed by shared global priors~\citep{basu21noregrets}. These models support few-shot adaptation and promote generalisation under limited supervision. Our non-parametric Bayesian approach extends this line by learning structured, data-driven priors informed by metadata without assuming strict inter-task similarities.
\cite{wan21mtts} considers context-dependent meta priors; however, they assume knowledge of the latent variables and functional relationships between context and reward generation and moreover, strong parametric assumptions about how latent variables are distributed.
\cite{hong2022hierbayesbandits} propose a hierarchical Bayesian model unifying the frameworks of meta multi task and federated bandits; however, their framework still needs specification for a prior distribution. 

\paragraph{Federated, Collaborative, and Structured Bandits.}
Federated and multi-agent bandits study decentralised settings where agents (such as users or devices) operate independently and synchronise occasionally~\citep{wu2016contextual, Huang2021fed_lin_context_bandits, shi21federated, huch2024rome}. These often assume clustered or homogeneous agent populations, using shared priors or latent factor models to coordinate learning. Our method removes such assumptions and allows task relationships to emerge organically via metadata-aware priors, enabling generalisation across unstructured and partially clustered regimes.

Collaborative filtering bandits treat rewards as a user-item matrix and exploit low-rank or clustered structure for generalisation~\citep{li2016collaborative}. Similarly, clustering-based methods like the Gang of Bandits~\citep{gentile2014online},\cite{ ban2024meta}, partition users into latent groups and pool observations to guide exploration. These approaches work well when clustering is a valid approximation. Later works soften this assumption by modelling probabilistic task similarities or considering misspecification of users~\citep{wang2023online}.

Another strand decomposes reward functions into global (shared) and local (task-specific) components, inspired by mixed-effect models~\citep{yang21impact,huch2024rome}. This additive decomposition captures both universal trends and task-specific nuances. Our approach shares this spirit but models these dependencies probabilistically using task metadata and hierarchical priors.

\paragraph{Cold-Start and Bayesian Optimisation Connections.}
Cold-start scenarios pose a major challenge in personalised systems, where new users or items lack sufficient historical data. Prior works address this via side information
~\citep{wang2017biucb}
, meta-learned priors~\citep{Bharadhwaj2019metalearningforcoldstart}, or population-level summaries~\citep{Alabdulrahman2019, shams2021cluster}. We extend these ideas by learning informative priors directly from task metadata, enabling effective adaptation without overfitting early sparse data.

Finally, Bayesian Optimisation (BO) has been extended to multi-task and meta-learning settings through conditional BO~\citep{chen2018bayesianoptimizationalphago, pearce2020practicalbayesianoptimizationobjectives}, which shares information across tasks via surrogate models. While our method is not BO-based, it aligns conceptually in using a hierarchical Bayesian approach to enable generalisation and structured knowledge transfer.

Across all these settings the unifying principle is the same: leveraging shared structure for efficient exploration and generalisation. Our method contributes to this landscape by providing a flexible, metadata-aware framework for modelling task heterogeneity without strong structural assumptions.

\section{Gaussian Process Modelling  Details}
\subsection{Gaussian Processes as Distributions over Functions}

Let \( \mathcal{X} \subset \mathbb{R}^d \) be a compact input domain. A Gaussian Process defines a distribution over real-valued functions \( f: \mathcal{X} \to \mathbb{R} \), such that for any finite collection of input points \( \{x_1, \dots, x_n\} \subset \mathcal{X} \), the associated function values \( (f(x_1), \dots, f(x_n)) \in \mathbb{R}^n \) follow a multivariate Gaussian distribution:
\[
f \sim \mathcal{GP}(m, k)
\quad \Longleftrightarrow \quad
(f(x_1), \dots, f(x_n)) \sim \mathcal{N}(\boldsymbol{m}, \boldsymbol{K}),
\]
where the mean vector \( \boldsymbol{m} \in \mathbb{R}^n \) and covariance matrix \( \boldsymbol{K} \in \mathbb{R}^{n \times n} \) are given by
\[
\boldsymbol{m}_i = m(x_i), \qquad \boldsymbol{K}_{ij} = k(x_i, x_j).
\]
The mean function \( m: \mathcal{X} \to \mathbb{R} \) and the kernel (covariance) function \( k: \mathcal{X} \times \mathcal{X} \to \mathbb{R} \) fully specify the GP. The kernel function must be symmetric and positive semi-definite, ensuring that for any finite set of inputs, the corresponding covariance matrix \( \boldsymbol{K} \) is positive semi-definite.

The choice of kernel controls the regularity properties of the sample paths of the GP. For example, the squared exponential kernel,
\[
k_{\text{SE}}(x, x') = \sigma_f^2 \exp\left(-\frac{\|x - x'\|^2}{2\ell^2}\right),
\]
yields smooth, infinitely differentiable functions. Matérn kernels allow more control over function roughness, depending on a smoothness parameter \( \nu \). For most standard kernels, sample paths are almost surely continuous.

Conceptually, GPs define probability measures over function spaces. While the reproducing kernel Hilbert space (RKHS) associated with \( k \) describes a subspace of functions with bounded norm, the support of the GP measure extends beyond the RKHS, including more general continuous functions.
\subsection{From Function Priors to Log-Densities}

To define a flexible prior over probability densities on a space \( \mathcal{Z} = \mathbb{R}^K \times \mathbb{R}^{d_{\text{obs}}} \), we place a GP prior on the logarithm of the density. Specifically, we let
\[
f \sim \mathcal{GP}(m, k), \quad f: \mathcal{Z} \to \mathbb{R},
\]
and define a density \( P^{(f)} \) on \( \mathcal{Z} \) via
\[
P^{(f)}(z) = \frac{\exp(f(z))}{Z_f}, \qquad Z_f = \int_{\mathcal{Z}} \exp(f(z)) \, dz.
\]
This approach induces a distribution over densities \( P^{(f)} \), where the normalisation constant \( Z_f \) ensures that \( P^{(f)} \) integrates to one. For common kernels and compact domains, the integral \( Z_f \) exists with high probability under the GP prior. When the domain is non-compact, it is often sufficient to ensure integrability by adding a negative quadratic drift or restricting \( f \) to decay appropriately at infinity.

This transformation of GP sample paths into densities via exponentiation and normalisation ensures that \( P^{(f)} \) is strictly positive and integrates to one. Since GPs generate continuous sample paths almost surely, the resulting densities are smooth and capable of modelling complex dependencies.

\subsection{Bayesian Nonparametric Density Modelling}

The prior \( Q(f) = \mathcal{GP}(m, k) \) induces a prior over densities \( Q(P^{(f)}) \). Each sample \( f \sim Q \) defines a unique joint distribution \( P^{(f)} \), and posterior inference over \( f \) yields a Bayesian posterior over densities. In our setting, this corresponds to learning a distribution over reward models \( P^{(f)}(\boldsymbol{\mu}_i, \boldsymbol{x}_{i,\text{obs}}) \), where both the arm mean vector \( \mu \in \mathbb{R}^K \) and observed context \( x_{\text{obs}} \in \mathbb{R}^{d_{\text{obs}}} \) are treated jointly.

Conditioned on interaction data \( \mathcal{H}_t \), we update the prior via Bayes' rule to obtain the posterior \( Q_t(f) = Q(f \mid \mathcal{H}_t) \). In turn, this defines posterior predictive distributions \( P^{(f)}(\mu \mid x_{\text{obs}}, \mathcal{H}_t) \), which are used to guide personalised decision-making. This modelling approach supports coherent uncertainty quantification at both the individual and population levels.

\subsection{Justification and Practical Use}

Modelling log-densities with GPs provides a principled way to perform density estimation without strong parametric assumptions. The transformation ensures that the resulting function is always a valid density function, and the GP prior allows expressing rich inductive biases through the choice of kernel.

Compared to kernel density estimators, log-GP models generalise better in higher dimensions, accommodate structured priors, and permit posterior inference under uncertainty. Compared to parametric latent variable models, log-GP densities are more robust to model misspecification and more flexible in capturing multi-modal or non-linear structure.

\subsection{Computational Challenges of GP-Based Density Modelling}

While Gaussian Processes (GPs) provide a flexible and nonparametric prior over log-densities, their practical application to density estimation is hampered by significant computational challenges.

First, to define a valid density from a function \( f \sim \mathcal{GP}(m, k) \), one must compute the normalisation constant
\[
Z_f = \int_{\mathcal{Z}} \exp(f(z)) \, dz,
\]
 This requires evaluating the function \( f \) over a discretised grid covering \( \mathcal{Z} \). In practice, we approximate \( Z_f \) using numerical quadrature or Monte Carlo integration, which involves computing \( f(z_j) \) for a dense set of grid points \( \{z_j\}_{j=1}^{M} \). This step alone incurs \( \mathcal{O}(M) \) cost per sample, where \( M \) grows exponentially with the input dimension due to the curse of dimensionality.

To obtain such evaluations, one must first draw a sample \( f \sim \mathcal{GP}(m, k) \). Since GPs are infinite-dimensional stochastic processes, sampling a full function is only possible on a finite set of evaluation points. Let \( \mathcal{Z}_M = \{z_j\}_{j=1}^M \subset \mathcal{Z} \) denote a grid of inputs. The restriction of \( f \) to \( \mathcal{Z}_M \) defines a multivariate Gaussian distribution:
\[
(f(z_1), \dots, f(z_M))^\top \sim \mathcal{N}(\boldsymbol{m}, \boldsymbol{K}),
\]
with \( \boldsymbol{m}_j = m(z_j) \) and \( \boldsymbol{K}_{ij} = k(z_i, z_j) \). Sampling from this distribution requires drawing a vector from a Gaussian with covariance matrix \( \boldsymbol{K} \in \mathbb{R}^{M \times M} \), which costs \( \mathcal{O}(M^3) \) for Cholesky decomposition and \( \mathcal{O}(M^2) \) for each sample.

Therefore, for each function sample:
\begin{itemize}
  \item Sampling the function over \( M \) points requires \( \mathcal{O}(M^3) \) preprocessing (or \( \mathcal{O}(M^2) \) with structured kernels or inducing points),
  \item Evaluating \( \exp(f(z)) \) over all \( z \in \mathcal{Z}_M \) incurs \( \mathcal{O}(M) \) cost,
  \item Approximating the integral \( Z_f \) adds another \( \mathcal{O}(M) \),
  \item This must be repeated for every sampled \( f \sim Q_t(f) \) during inference or planning.
\end{itemize}

More fundamentally, Bayesian inference in our model requires marginalising over all such functions. Letting \( Q(f) = \mathcal{GP}(m, k) \), the posterior given history \( \mathcal{H}_t \) is
\[
Q(f \mid \mathcal{H}_t) \propto P(\mathcal{H}_t \mid f) \cdot Q(f),
\]
and predictions require evaluating expectations of the form
\[
\mathbb{E}_{f \sim Q(f \mid \mathcal{H}_t)}\left[\Phi(f)\right],
\]
for functionals \( \Phi(f) \) such as expected rewards or optimal arm probabilities.

Since this outer integral is over an infinite-dimensional space of functions, exact inference is intractable. The combination of:
\begin{enumerate}
  \item integration over \( \mathcal{Z} \) for each function sample to evaluate likelihoods,
  \item integration over function space for Bayesian inference,
\end{enumerate}
renders exact computation computationally prohibitive, particularly in online or sequential settings such as bandits.

To overcome these issues, we employ a truncated Karhunen–Loève expansion of the GP prior, projecting \( f \) into a low-dimensional linear span of orthonormal eigenfunctions:
\[
f(z) \approx \sum_{\ell=1}^{L} \theta_\ell \phi_\ell(z), \quad \theta_\ell \sim \mathcal{N}(0, \lambda_\ell),
\]
where \( \{(\lambda_\ell, \phi_\ell)\} \) are the leading eigenpairs of the covariance operator induced by \( k \). We perform posterior inference over the finite-dimensional coefficient vector \( \boldsymbol{\theta} \in \mathbb{R}^L \) using Sequential Monte Carlo (SMC), which allows efficient sampling and updating over time. This approximation enables scalable inference while preserving the expressive modelling capacity of the full GP prior. Further implementation details are provided in Appendix~\ref{appendix:smc}.

\subsection{Karhunen–Loève Expansion and Particle Approximation}
\label{appendix:kl-smc}

To make inference over Gaussian Process-defined log-densities computationally feasible, we adopt a finite-dimensional approximation based on the Karhunen–Loève (KL) expansion. This enables us to project the infinite-dimensional function space of the GP onto a lower-dimensional basis of orthonormal eigenfunctions, and reduces the problem of sampling functions to sampling standard Gaussian vectors in Euclidean space.

Formally, given a zero-mean Gaussian Process \( f \sim \mathcal{GP}(0, k) \) defined over a compact domain \( \mathcal{Z} \subset \mathbb{R}^{K + d_{\text{obs}}} \), the KL expansion of \( f \) takes the form:
\[
f(z) = \sum_{m=1}^\infty \sqrt{\lambda_m} \, \xi_m \, \phi_m(z),
\]
where \( \{\phi_m\}_{m=1}^\infty \) are the orthonormal eigenfunctions of the covariance operator associated with \( k \), and \( \{\lambda_m\}_{m=1}^\infty \) are the corresponding eigenvalues. The coefficients \( \{\xi_m\} \) are independent standard normal random variables, \( \xi_m \sim \mathcal{N}(0, 1) \).

In practice, we truncate this expansion to the top \( M \) components:
\[
f(z) \approx \sum_{m=1}^M \sqrt{\lambda_m} \, \xi_m \, \phi_m(z),
\]
yielding a finite-dimensional approximation of the function space. The GP prior thus induces a distribution over the coefficients \( \boldsymbol{\xi} = (\xi_1, \dots, \xi_M) \), which now parameterise the function \( f \).

To numerically construct the empirical KL basis, we discretise \( \mathcal{Z} \) with a grid of points \( \{z_\ell = (\boldsymbol{\mu}_\ell, \boldsymbol{x}_\ell)\}_{\ell=1}^L \), and compute the kernel matrix
\[
K_{\ell \ell'} = k(z_\ell, z_{\ell'}).
\]
Performing eigendecomposition \( K = \Phi \Lambda \Phi^\top \) gives us the top \( M \) eigenvalues \( \lambda_1, \dots, \lambda_M \) and the corresponding empirical eigenvectors \( \phi_m \in \mathbb{R}^L \), which approximate the continuous eigenfunctions at the grid points.

Each function sample \( f^{(j)} \) can then be constructed by drawing a particle \( \boldsymbol{\xi}^{(j)} \sim \mathcal{N}(0, I_M) \) and computing:
\[
f^{(j)}(z) = \sum_{m=1}^M \sqrt{\lambda_m} \, \xi^{(j)}_m \, \phi_m(z).
\]
This representation allows us to sample functions from the GP by sampling low-dimensional Gaussian vectors. Since all dependence on \( z \) is encoded in the basis functions \( \phi_m(z) \), this approximation decouples function complexity from inference dimensionality.

\paragraph{Particle Approximation and Inference.} Using this KL representation, we can perform Bayesian inference over log-densities by maintaining a set of particles \( \{\boldsymbol{\xi}^{(j)}\}_{j=1}^N \), each corresponding to a log-density sample \( f^{(j)} \). The induced density \( P^{(f^{(j)})} \) is obtained by exponentiating and normalising:
\[
P^{(f^{(j)})}(z) = \frac{\exp(f^{(j)}(z))}{\int \exp(f^{(j)}(z')) dz'}.
\]
These densities are then used to compute likelihoods, update particle weights, and propagate posterior beliefs over time using Sequential Monte Carlo (SMC). The particle-based KL approximation thus enables tractable Bayesian learning over densities, supporting dynamic posterior updates as new interaction data arrives.

\paragraph{Solving the Inverse Problem.} In the context of density estimation, the fundamental inverse problem is to infer a function \( f \) such that \( \exp(f(z)) / Z_f \) best explains the observed data under a generative model. The KL expansion transforms this into a search over a finite-dimensional coefficient space \( \mathbb{R}^M \), making posterior inference feasible. By optimising or sampling over \( \boldsymbol{\xi} \in \mathbb{R}^M \), we indirectly explore the space of plausible log-densities, while avoiding the intractability of directly estimating infinite-dimensional functionals.

\paragraph{Computational Efficiency.} Once the kernel eigendecomposition is computed (costing \( \mathcal{O}(L^3) \), where \( L \) is the number of grid points), each function sample requires \( \mathcal{O}(M L) \) operations to evaluate over the grid. This makes online particle filtering feasible even in moderately high-dimensional spaces, and enables repeated sampling and evaluation during acquisition and planning steps.

We find that retaining the top \( M \approx 50\)--\(100 \) components typically suffices to capture most functional variability in practice, especially under smooth kernels. The combination of KL approximation and particle-based Bayesian updates enables efficient and expressive log-density estimation in our multi-task bandit setting.

\subsection{Sequential Monte Carlo for Bayesian Inference in Multi-Task Multi-Armed Bandits}
\label{appendix:smc}

In our setting of multi-task, multi-armed bandits, users arrive sequentially, each associated with a new task defined by an unknown reward model. We aim to model the underlying heterogeneity across users using a Bayesian nonparametric prior over log-densities, implemented through a Gaussian Process (GP) over the joint space \( \mathcal{Z} = \mathbb{R}^K \times \mathbb{R}^{d_{\text{obs}}} \). As new interaction data becomes available, our posterior over latent reward models must be updated efficiently and scalably in real time. To this end, we employ \emph{Sequential Monte Carlo} (SMC), a family of algorithms designed to perform approximate Bayesian inference in dynamic and computationally intensive settings.

\paragraph{Primer on SMC.} SMC approximates a sequence of posterior distributions \( \{\pi_t(\theta)\}_{t=1}^T \) using a collection of weighted particles \( \{ \theta^{(i)}_t, w^{(i)}_t \}_{i=1}^N \), where each \( \theta^{(i)}_t \in \mathbb{R}^M \) is a sample from the parameter space (in our case, the KL coefficients representing a sample \( f^{(i)} \) from the GP), and \( w^{(i)}_t \) is its associated importance weight. These particles are updated as new data arrives, using importance sampling, resampling, and rejuvenation.

\paragraph{Particle representation.} Using the Karhunen–Loève (KL) expansion described in Appendix~\ref{appendix:kl-smc}, we express each GP sample \( f^{(i)} \) as
\[
f^{(i)}(z) = \sum_{m=1}^M \sqrt{\lambda_m} \, \xi^{(i)}_m \, \phi_m(z),
\]
where \( \boldsymbol{\xi}^{(i)} \in \mathbb{R}^M \) are standard Gaussian coefficients. These KL coefficients form the latent variables \( \theta^{(i)}_t \) over which inference is performed.

\paragraph{Importance weighting.} When a new observation \( (\boldsymbol{x}_{i,\text{obs}}, a_t, r_t) \) is made, the particles are reweighted according to the likelihood of the observed reward under the density induced by each particle:
\[
w_t^{(i)} \propto w_{t-1}^{(i)} \cdot P^{(f^{(i)})}(r_t \mid a_t, \boldsymbol{x}_{i,\text{obs}}, \mathcal{H}_{i(t),t}),
\]
with normalisation to ensure \( \sum_i w_t^{(i)} = 1 \). This updates the approximation of the posterior to reflect the new information.

\paragraph{Resampling.} As more data accumulates, particle degeneracy can occur, where most particles have negligible weight. To address this, we monitor the effective sample size (ESS),
\[
\text{ESS} = \left( \sum_{i=1}^N (w_t^{(i)})^2 \right)^{-1},
\]
and trigger a resampling step if \( \text{ESS} < \tau N \), for some threshold \( \tau \in (0,1) \). Resampling replicates high-weight particles and discards low-weight ones, ensuring that computational effort is focused on promising regions of the parameter space.

\paragraph{Rejuvenation via Langevin dynamics.} Resampling reduces diversity in the particle population, as duplicates of the same particle are created. To mitigate this, we apply a \emph{rejuvenation} step using MCMC. Rather than relying on uninformed random-walk proposals, we employ Langevin dynamics, which leverages gradient information:
\[
\theta' = \theta + \frac{\eta}{2} \nabla_\theta \log \pi_t(\theta) + \sqrt{\eta} \, \epsilon, \qquad \epsilon \sim \mathcal{N}(0, I),
\]
where \( \pi_t(\theta) \propto p(\theta) \cdot p(\mathcal{H}_t \mid \theta) \) is the current posterior, and \( \eta \) is a step size. The proposal is accepted or rejected using a Metropolis–Hastings rule to correct for discretisation error. This Langevin-based rejuvenation allows for efficient exploration of high-dimensional posteriors and makes use of the smooth structure induced by the GP prior.

\paragraph{Tempered transitions.} To further improve robustness (particularly in multi-modal or sharply peaked posterior landscapes) we introduce a sequence of intermediate target distributions using a temperature parameter \( \beta \in [0,1] \):
\[
\pi_\beta(\theta) \propto p(\theta) \cdot p(\mathcal{H}_t \mid \theta)^\beta.
\]
By gradually increasing \( \beta \) from 0 to 1, particles adapt incrementally from the prior to the full posterior. This \emph{tempering} scheme helps maintain stability and avoids early particle collapse.

\paragraph{Why SMC over MCMC.} Although the underlying generative model \( f \sim \mathcal{GP}(0, k) \) is fixed across time, our posterior belief over \( f \) evolves as new data is observed. Standard Markov Chain Monte Carlo (MCMC) methods, such as Hamiltonian Monte Carlo, are designed for sampling from a fixed posterior distribution. They typically require restarting the inference procedure from scratch each time the data changes entailing significant computational overhead and poor scalability in online settings. In contrast, SMC naturally accommodates sequential updates and avoids full recomputation, making it ideal for interactive systems where observations arrive incrementally.

\paragraph{Computational complexity.} The main computational costs in SMC arise from three steps:
\begin{itemize}
  \item \textbf{Likelihood evaluation:} For each particle \( \theta^{(i)} \), evaluating \( f^{(i)}(z) \) over a grid of size \( L \) costs \( \mathcal{O}(M L) \).
  \item \textbf{Weight update:} Computing log-likelihoods for new observations across \( N \) particles takes \( \mathcal{O}(N M L) \) per time step.
  \item \textbf{Langevin MCMC:} Each rejuvenation step requires computing gradients and log-densities, typically \( \mathcal{O}(M L) \) per particle per step.
\end{itemize}
The initial eigendecomposition of the kernel matrix over the grid (to construct KL components) has one-time cost \( \mathcal{O}(L^3) \), but is shared across all users and reused throughout.

 Sequential Monte Carlo provides a principled, flexible, and scalable method for posterior inference in our nonparametric multi-task bandit model. It accommodates functional priors, supports online updates, and integrates uncertainty into planning. Through importance sampling, adaptive resampling, Langevin-based rejuvenation, and tempering, SMC enables robust approximation of complex posterior distributions in dynamic decision-making environments.

 \subsection{Conditioning on User History via Particle Weighting}
\label{appendix:user-conditioning}

Given a Gaussian Process prior over log-densities \( f(\boldsymbol{\mu}, \boldsymbol{x}_{\text{obs}}) \sim \mathcal{GP}(m, k) \), and a user-specific task context \( \boldsymbol{x}_{\text{obs}} \), we wish to infer a posterior distribution over arm mean vectors \( \boldsymbol{\mu} \in \mathbb{R}^K \) using a user's reward history \( \mathcal{H}_j = \{(a_i, r_i)\}_{i=1}^n \).

\paragraph{Sampling from the conditional log-density.}  
For a fixed \( \boldsymbol{x}_{\text{obs}} \), and each function sample \( f^{(s)} \) drawn via the KL approximation (see Appendix~\ref{appendix:kl-smc}), we evaluate it on a discretised grid \( \{\boldsymbol{\mu}_\ell\}_{\ell=1}^L \subset \mathbb{R}^K \). This gives:
\[
f^{(s)}_\ell := f^{(s)}(\boldsymbol{\mu}_\ell, \boldsymbol{x}_{\text{obs}}),
\]
which defines an unnormalised log-density over arm means. We convert this to a discrete probability distribution via a softmax:
\[
\pi^{(s)}_\ell = \frac{\exp(f^{(s)}_\ell)}{\sum_{\ell'=1}^L \exp(f^{(s)}_{\ell'})},
\]
yielding a discrete approximation to the conditional density \( \pi^{(s)}(\boldsymbol{\mu} \mid \boldsymbol{x}_{\text{obs}}) \). We then draw samples \( \boldsymbol{\mu}^{(s)} \sim \pi^{(s)} \) from this distribution, yielding a collection of candidate mean vectors.

\paragraph{Likelihood-based particle weighting.}  
Each user provides a sequence of reward observations \( \mathcal{H}_j = \{(a_i, r_i)\}_{i=1}^n \), with rewards modeled as:
\[
r_i \sim \mathcal{N}(\mu_{a_i}, \sigma^2),
\]
where \( \mu_{a_i} \) denotes the mean reward of arm \( a_i \) for this user. For each sampled mean vector \( \boldsymbol{\mu}^{(s)} \), we compute its likelihood under the observed rewards:
\[
\mathcal{L}^{(s)} := \prod_{i=1}^n \mathcal{N}(r_i \mid \mu^{(s)}_{a_i}, \sigma^2).
\]
This likelihood is interpreted as an unnormalised importance weight for the sample \( \boldsymbol{\mu}^{(s)} \). After normalisation, we obtain the particle weights:
\(
w^{(s)} := \frac{\mathcal{L}^{(s)}}{\sum_{s'=1}^N \mathcal{L}^{(s')}}.
\)
These weights define an empirical approximation to the posterior \( p(\boldsymbol{\mu} \mid \mathcal{H}_j, \boldsymbol{x}_{\text{obs}}) \), computed over the grid based on the KL expansion.

\paragraph{Approximate marginal likelihood as convolution.}  
This particle-based conditioning can be understood as a discrete approximation to the convolution integral:
\(
p(\mathcal{H}_j \mid \pi) = \int_{\mathbb{R}^K} \left[\prod_{i=1}^n \mathcal{N}(r_i \mid \mu_{a_i}, \sigma^2)\right] \pi(\boldsymbol{\mu} \mid \boldsymbol{x}_{\text{obs}}) \, d\boldsymbol{\mu}.
\)
This integral arises in marginal likelihood evaluation or evidence computation but is generally intractable for arbitrary \( \pi \). Instead, we approximate it using samples \( \boldsymbol{\mu}^{(s)} \sim \pi^{(s)} \) and compute the marginal likelihood as:
\(
\widehat{p}(\mathcal{H}_j \mid \pi^{(s)}) \approx \frac{1}{N} \sum_{s=1}^N \mathcal{L}^{(s)}.
\)
Thus, our particle-weighted approximation captures both the posterior over arm means and a tractable estimate of the likelihood required for Bayesian updating of the latent log-density model.

\begin{algorithm}[t!]
\caption{SMC Inference for GP-Based Log-Density Model}
\label{alg:smc-refined}
\begin{algorithmic}[1]
\Require GP kernel \( k \); grid \( \{z_\ell\}_{\ell=1}^L \); KL truncation \( M \); particles \( N \); step size \( \eta \); ESS threshold \( \tau \); max SMC updates \( n_{\text{max}} \); noise \( \sigma^2 \)

\Statex \Comment{KL Basis Construction}
\State Compute kernel matrix \( K_{\ell \ell'} = k(z_\ell, z_{\ell'}) \in \mathbb{R}^{L \times L} \)
\State Eigendecompose \( K = \Phi \Lambda \Phi^\top \), retain top \( M \) eigenpairs

\Statex \Comment{Initialise Particles}
\For{$i = 1$ to $N$}
  \State Sample \( \boldsymbol{\xi}^{(i)} \sim \mathcal{N}(0, I_M) \)
  \State Set \( f^{(i)}(z_\ell) = \sum_{m=1}^M \sqrt{\lambda_m} \, \xi_m^{(i)} \, \phi_m(\ell) \)
  \State Set \( w^{(i)} \gets 1/N \)
\EndFor

\For{each round $t = 1, 2, \dots$}
  \State Observe context \( \boldsymbol{x}_{k,\text{obs}} \), reward \( r_t \), and user history \( \mathcal{H}_{k,t} \)

  \State \textbf{Repeat} up to \( n_{\text{max}} \) times or until \( \text{ESS} \geq \tau N \):
  \Statex \hspace{1em} \Comment{Importance Weighting}
  \For{$i = 1$ to $N$}
    \State Evaluate GP sample \( f^{(i)}(\boldsymbol{\mu}_\ell, \boldsymbol{x}_{k,\text{obs}}) \) over grid
    \State Compute softmax: \( \pi^{(i)}_\ell = \exp(f^{(i)}_\ell) / \sum_{\ell'} \exp(f^{(i)}_{\ell'}) \)
    \State Sample \( \boldsymbol{\mu}^{(i)} \sim \pi^{(i)} \)
    \State Compute likelihood:
    \[
    \mathcal{L}^{(i)} = \prod_{(a_j, r_j) \in \mathcal{H}_{i,t}} \mathcal{N}(r_j \mid \mu^{(i)}_{a_j}, \sigma^2)
    \]
    \State Update \( w^{(i)} \gets w^{(i)} \cdot \mathcal{L}^{(i)} \)
  \EndFor
  \State Normalize: \( w^{(i)} \gets w^{(i)} / \sum_j w^{(j)} \)
  \State Compute \( \text{ESS} = \left(\sum_i (w^{(i)})^2\right)^{-1} \)

  \If{\( \text{ESS} < \tau N \)}
    \State \Comment{Resample and Rejuvenate}
    \State Resample \( \{ \tilde{\boldsymbol{\xi}}^{(i)} \} \sim \{ w^{(i)} \} \), reset \( w^{(i)} \gets 1/N \)
    \For{$i = 1$ to $N$}
      \State Compute \( \nabla_{\xi} \log p(\mathcal{H}_{i,t} \mid \tilde{\boldsymbol{\xi}}^{(i)}) \)
      \State Propose Langevin step:
      \[
      \boldsymbol{\xi}'^{(i)} = \tilde{\boldsymbol{\xi}}^{(i)} + \tfrac{\eta}{2} \nabla \log p + \sqrt{\eta} \, \epsilon, \quad \epsilon \sim \mathcal{N}(0, I)
      \]
      \State Accept \( \boldsymbol{\xi}' \) via Metropolis–Hastings
    \EndFor
  \Else
    \State \textbf{Break}
  \EndIf
  \State \textbf{End Repeat}
\EndFor
\end{algorithmic}
\end{algorithm}

\begin{figure}[t!]
    \centering
    \includegraphics[width=\linewidth]{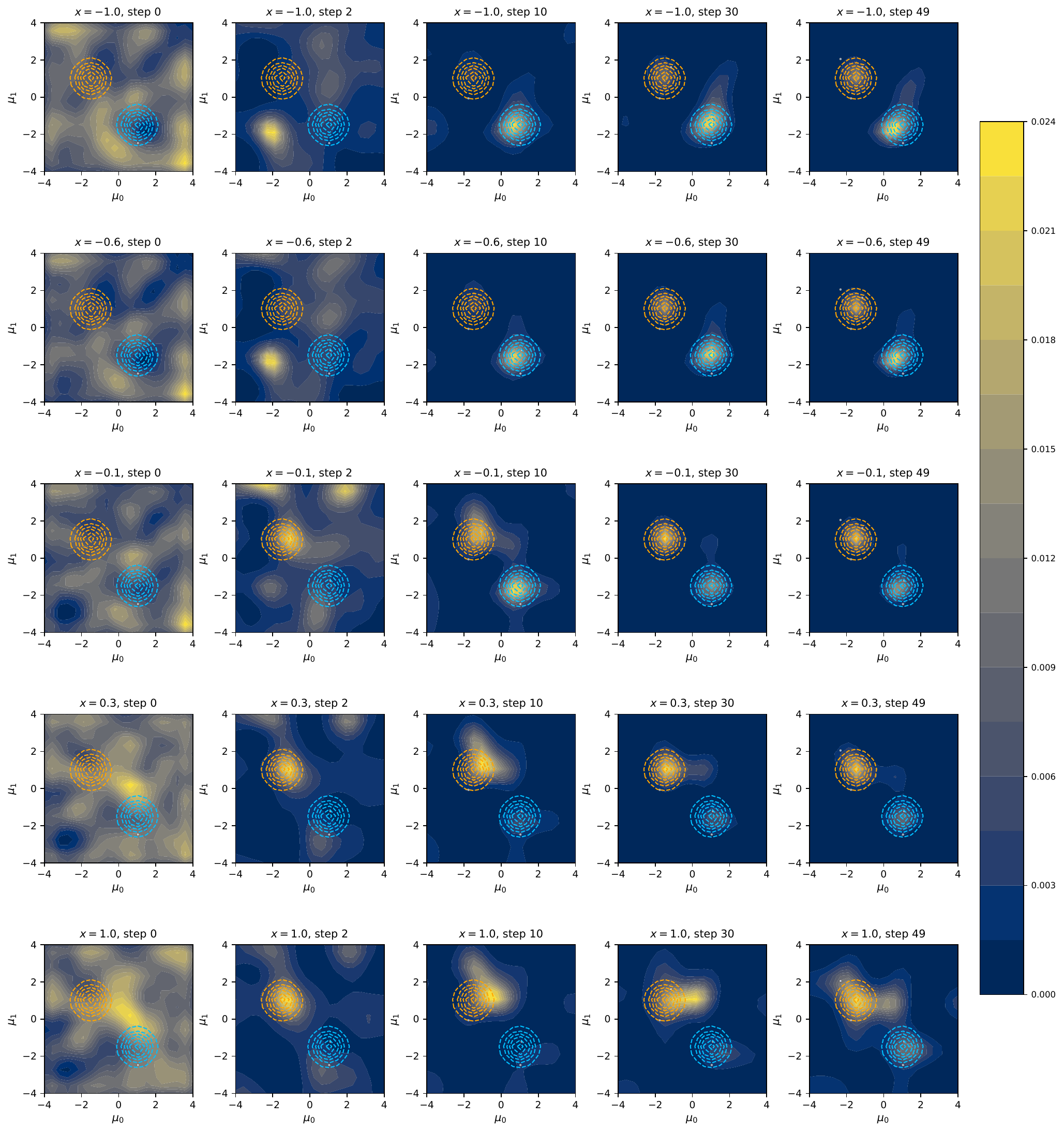}
    \caption{Sequential approximation of a context-dependent mixture of Gaussians using our KL-expanded SMC-GP framework. 
        The underlying target is a mixture of two Gaussians with context-dependent weights: for each context \( x \in [-1,1] \), the mixture weight is given by \( \sigma(x) = \frac{1}{1 + \exp(-x)} \). 
        We visualise the marginal distributions over the mean rewards \( \mu_0 \) and \( \mu_1 \) at five contexts and five time steps \( t \in \{0, 2, 10, 30, 49\} \).
        At \( t=0 \), particles are drawn from the GP prior; as data accumulates, the posterior increasingly concentrates around the true multimodal density. }
    \label{fig:mets_regret_partial_context}
\end{figure}

\subsection{Implementation of Thompson Sampling via Particle-Based Posterior}
\label{appendix:ts-details}

We implement Thompson Sampling using our KL-truncated Gaussian Process model, which defines a log-density over arm mean vectors \( \boldsymbol{\mu} \in \mathbb{R}^K \), conditioned on user context \( \boldsymbol{x}_{i,\text{obs}} \) and interaction history \( \mathcal{H}_{i,t} \). The posterior over log-densities is approximated by a set of functional particles \( \{f^{(s)}, w^{(s)}\}_{s=1}^N \).

At each decision time \( t \), the algorithm proceeds as follows:

\begin{enumerate}
    \item \textbf{Sample a functional particle.}  
    Draw \( s \sim \text{Categorical}(w^{(1)}, \dots, w^{(N)}) \), selecting a GP sample \( f^{(s)} \).
    
    \item \textbf{Evaluate log-density on grid.}  
    For a discretised grid \( \{ \boldsymbol{\mu}_\ell \}_{\ell=1}^L \subset \mathbb{R}^K \), compute:
    \[
    f^{(s)}_\ell := f^{(s)}(\boldsymbol{\mu}_\ell, \boldsymbol{x}_{i,\text{obs}}),
    \quad
    \pi^{(s)}_\ell = \frac{\exp(f^{(s)}_\ell)}{\sum_{\ell'} \exp(f^{(s)}_{\ell'})}.
    \]

    \item \textbf{Compute arm optimality distribution.}  
    For each arm \( a \in \{1, \dots, K\} \), define:
    \[
    \hat{p}_a^{(s)} := \sum_{\ell \in \mathcal{I}_a} \pi^{(s)}_\ell, \quad \text{where } \mathcal{I}_a = \left\{ \ell : a = \arg\max_{a'} \mu_{\ell,a'} \right\}.
    \]

    \item \textbf{Sample the action.}  
    Select \( a_{i,t} \sim \text{Categorical}(\hat{p}_1^{(s)}, \dots, \hat{p}_K^{(s)}) \).
\end{enumerate}

This implementation retains the fully Bayesian structure of Thompson Sampling by sampling from the posterior over log-densities, inducing a predictive distribution over optimal actions through marginalisation over mean vectors. It captures uncertainty both over function space and over arm-level optima without relying on parametric assumptions.

\subsection{Global IDS with Particle Weight Entropy}
\label{appendix:gids-entropy}

We provide a practical implementation of Global Information Directed Sampling (GIDS) using entropy over the particle weights to quantify information gain. This approach avoids recomputing full posterior updates by approximating how the particle weights \( \{ w^{(s)} \} \) would change under hypothetical observations. The key idea is to evaluate, for each arm, the expected entropy reduction in the posterior over functional samples \( f \sim Q_t(f) \), represented by the particles.

\paragraph{Setup.} 
We begin by sampling a single function particle \( f^{(s)} \sim Q_t(f) \) according to the normalized weights \( w^{(s)} \). Given the user context \( \boldsymbol{x}_{i,\text{obs}} \) and history \( \mathcal{H}_{i,t} \), we evaluate the log-density of \( f^{(s)} \) over the grid \( \{\boldsymbol{\mu}_\ell\}_{\ell=1}^L \), and compute the corresponding softmax-normalized density \( \pi^{(s)}_\ell \), as described in Appendix~\ref{appendix:ts-details}.

\paragraph{Regret estimation.}
For each arm \( a \in \{1, \dots, K\} \), we compute the expected regret under this log-density:
\[
\text{Regret}_t^{(i,f)}(a) = \sum_{\ell=1}^L \pi^{(s)}_\ell \left( \max_{a'} \mu_{\ell,a'} - \mu_{\ell,a} \right).
\]

\paragraph{Entropy-based information gain.}
We define the uncertainty over \( f \sim Q_t(f) \) by the entropy of the normalized particle weights:
\[
H(Q_t(f)) := - \sum_{s=1}^N w^{(s)} \log w^{(s)}.
\]
For each arm \( a \), and for a range of plausible reward outcomes \( r \in \mathcal{R} \), we evaluate the updated particle weights assuming the reward \( r \) is observed from pulling arm \( a \). Specifically, we recompute each particle's likelihood:
\[
\mathcal{L}^{(s')}(r \mid a) = \sum_{\ell=1}^L \pi^{(s')}_\ell \, \mathcal{N}(r \mid \mu_{\ell,a}, \sigma^2),
\]
and update weights as:
\[
\tilde{w}^{(s')} \propto w^{(s')} \cdot \mathcal{L}^{(s')}(r \mid a).
\]
Normalizing these updated weights yields a new entropy estimate \( H(Q_t(f \mid r, a)) \). The expected information gain for arm \( a \) is then:
\[
\text{EIG}_t^{(i,f)}(a) = H(Q_t(f)) - \sum_{r \in \mathcal{R}} P(r \mid a, f^{(s)}) \cdot H(Q_t(f \mid r, a)),
\]
where the marginal \( P(r \mid a, f^{(s)}) = \sum_{\ell=1}^L \pi^{(s)}_\ell \, \mathcal{N}(r \mid \mu_{\ell,a}, \sigma^2) \).

\paragraph{Policy optimization.}
We now optimize the GIDS objective over the probability simplex:
\[
\pi_f^* = \arg\min_{\pi \in \Delta_K} \frac{\left( \sum_a \pi_a \, \text{Regret}_t^{(i,f)}(a) \right)^2}{\sum_a \pi_a \, \text{EIG}_t^{(i,f)}(a) + \epsilon}.
\]
This is solved via softmax-normalized exponentiated gradient descent over \( \Delta_K \), using log-scores \( \log \pi_a \propto -\Psi_a \), where \( \Psi_a \) is the GIDS acquisition score for arm \( a \).

\paragraph{Action selection.}
Finally, the action \( a_{i,t} \sim \pi_f^* \) is sampled from the optimized GIDS policy for the sampled particle \( f^{(s)} \). This completes the policy loop without averaging over particles, consistent with our goal of reducing structural uncertainty.

This approach allows approximate yet scalable optimization of a meta-level acquisition function under nonparametric structural uncertainty.

\section{Experimental Details and Environments}
\label{appendix:experiments}

We provide full specifications of the synthetic environments used in our experiments. These were designed to evaluate generalisation across tasks, uncertainty modelling under latent structure, and robustness to model misspecification.

\paragraph{General Setup.} Each environment is repeated over 10 random seeds. Unless otherwise stated, we use \( K = 3 \) arms and $1$-dimensional observable context. Observation noise is set to \( \sigma = 0.1 \). 

\subsection{Environment 1: Mixture of Gaussians with Context-Dependent Modes}
\label{app:mog}

This environment tests the agent's ability to infer and adapt to context-dependent subpopulation structure using a mixture of Gaussians as the prior over arm rewards. The task distribution is multimodal, with the dominant mode depending on the user’s observed context.

\paragraph{Context Space.} Each user is associated with a scalar context \( \boldsymbol{x}_\text{obs}\in [-1, 1] \), sampled uniformly. This context controls both the mixture weights and the reward variance.

\paragraph{Reward Distribution.} Given context \( \boldsymbol{x}_\text{obs} \), the true latent reward vector \( \boldsymbol{\mu} \in \mathbb{R}^3 \) for the \( K = 3 \) arms is drawn from a mixture of two Gaussians:
\[
P^*(\boldsymbol{\mu} \mid \boldsymbol{x}_\text{obs}) = \sum_{j=1}^2 v_j(\boldsymbol{x}_\text{obs}) \cdot \mathcal{N}(\boldsymbol{\mu} \mid \boldsymbol{\mu}^{(j)}(\boldsymbol{x}_\text{obs}), \Sigma^{(j)}),
\]
where the mixing coefficient is dependent on the context:
\[
v_1(\boldsymbol{x}_\text{obs}) = \frac{1}{1 + e^{-\boldsymbol{x}_\text{obs}}}, ,\qquad v_2(\boldsymbol{x}_\text{obs}) =1 -v_1(\boldsymbol{\boldsymbol{x}_\text{obs}})
\]

\paragraph{Mixture Means.} The two component means are fixed as:
\[
\boldsymbol{\mu}^{(1)} = [1.8,\; 1.0,\; -1.0], \qquad \boldsymbol{\mu}^{(2)} = [1.0,\; 1.9,\; -1.8].
\]
Each mode induces a different arm ranking, thereby encoding distinct treatment subgroups.

\paragraph{Context-Dependent Variance.} The variance of both components depends linearly on the context:
\[
\Sigma^{(j)}(\boldsymbol{x}_\text{obs}) = (0.2 + 0.1\boldsymbol{x}_\text{obs})^2 \cdot I_3, \quad j \in \{1, 2\},
\]
producing higher uncertainty in rewards for users with higher context values.

\paragraph{Observation Model.} When arm \( a \) is pulled, the observed reward is sampled as:
\[
r \sim \mathcal{N}(\mu_a, \sigma^2), \qquad \text{with } \sigma^2 = 0.01.
\]

\paragraph{Recruitment Scheme.} We simulate 30 recruitment rounds, each introducing a batch of 5 new users (150 users total). Every user interacts with the system over 5 decision steps before exiting. This setup mimics episodic deployments in real-world personalized recommendation or healthcare settings, where patients enter in cohorts and interact over a short duration.

\subsection{Environment 2: Partially Observed Linear Contextual Bandit}
\label{app:partial-linear}

This environment evaluates the agent’s ability to reason under uncertainty when only partial information about the user’s context is available. The reward structure is linear in the full context vector, but only a single observed component is provided to the algorithm.

\paragraph{Context Structure.} Each user is assigned a 4-dimensional context vector:
\[
\boldsymbol{x} = [x_o,\; x_c,\; x_d,\; 1],
\]
where:
\begin{itemize}
    \item \( x_o \sim \mathcal{U}(-1, 1) \) is an \textbf{observed} continuous feature,
    \item \( x_c \sim \mathcal{N}(0.5, 0.1^2) \) is a \textbf{latent} continuous feature,
    \item \( x_d \sim \text{Bernoulli}(\sigma(x_o)) \) is a \textbf{latent} binary feature with probability controlled by the observed context,
    \item \( 1 \) is a fixed bias term.
\end{itemize}
Only the scalar \( x_o \) is revealed to the agent; the remaining dimensions of \( \boldsymbol{x} \) are unobserved.

\paragraph{Reward Model.} The mean reward for each arm \( a \in \{0, 1, 2\} \) is a fixed linear function of the full context:
\[
\mu_a = \boldsymbol{w}_a^\top \boldsymbol{x}, \qquad r_{i,t} \sim \mathcal{N}(\mu_a, \sigma^2),
\]
where the noise variance is set to \( \sigma^2 = 0.01 \). The weight matrix is fixed as:
\[
W =
\begin{bmatrix}
0.6 & 0.1 & 1.0 & -0.9 \\
0.3 & 0.3 & -1.0 & 0.9 \\
0.1 & -0.2 & -0.3 & 0.1
\end{bmatrix}.
\]

\paragraph{Learning Challenge.} Since the agent observes only \( x_o \), but the reward depends nontrivially on latent variables \( x_c \) and \( x_d \), a purely linear policy over \( x_o \) is insufficient. Effective inference requires marginalizing over plausible latent configurations. This tests the model’s ability to represent and reason over latent heterogeneity in the context.

\paragraph{Recruitment Protocol.} We adopt the same episodic structure as in Environment~\ref{app:mog}, with 30 rounds of recruitment, 5 new users per round, and each user interacting over 5 time steps. This episodic design simulates real-world deployments where users engage in short sessions.

This environment follows the Gaussian bandit setting proposed by \citet{wan21mtts}, using a Bayesian linear mixed model (LMM) with both fixed and random effects. It is used to evaluate model performance under both aligned and misaligned structural assumptions.

\subsection{Environment 3: Bayesian Linear Mixed Model (LMM)}
\label{app:lmm}

This reproduces the Gaussian bandit setting of \citet{kveton2021meta} using a fixed-effects + random-effects decomposition.

\paragraph{Context and Feature Map.}  
Each user is associated with a scalar metadata feature \( \boldsymbol{x}_i \sim \mathcal{U}(-1,1) \). We define a fixed, known linear mapping \( \phi(\boldsymbol{x}, a) \in \mathbb{R}^2 \) from the context–arm pair to features:
\[
\phi(\boldsymbol{x}, a) = [\rho_a \cdot \boldsymbol{x}, 1],
\]
where the slope coefficients are fixed as:
\[
\rho = [0.9,\ -1.1,\ 0.2] \quad \text{for arms } a = 1, 2, 3.
\]
This yields an input dimension \( d = 2 \) per arm. For user \( i \), the full arm-level design matrix is:
\[
\Phi_i =
\begin{bmatrix}
0.9 \cdot \boldsymbol{x}_i & 1 \\
-1.1 \cdot \boldsymbol{x}_i & 1 \\
0.2 \cdot \boldsymbol{x}_i & 1
\end{bmatrix}
\in \mathbb{R}^{3 \times 2}.
\]

\paragraph{Fixed and Random Effects.}  
Latent reward means for user \( i \) are given by:
\[
\boldsymbol{r}_i = \Phi_i^\top\theta + \delta_i,
\]
where:
\[
\theta \sim \mathcal{N}(\mu_\theta, \Sigma_\theta), \quad \mu_\theta = \mathbf{1}_2, \quad \Sigma_\theta = 0.25 \cdot I_2,
\]
\[
\delta_i \sim \mathcal{N}(0, \Sigma), \quad \Sigma = 0.05^2 \cdot I_3.
\]
The fixed-effect vector \( \theta \) captures shared structure across all users and arms, while \( \delta_i \) models user-specific deviations, allowing heterogeneity unaccounted for by the metadata.

\paragraph{Misaligned Setting.}  
To test robustness to model misspecification, we define a prior over \( \theta \) using a bimodal Gaussian mixture:
\[
\theta \sim 
\begin{cases}
\mathcal{N}(-\mathbf{1}_2, 0.25 \cdot I_2), & \text{with probability } 0.5, \\
\mathcal{N}(+\mathbf{1}_2, 0.25 \cdot I_2), & \text{with probability } 0.5.
\end{cases}
\]
This introduces structural ambiguity at the population level, challenging parametric methods such as \textsc{MTTS}, while remaining tractable for our nonparametric GP-based model.

\paragraph{Recruitment Schemes.}  
We adopt distinct user recruitment protocols depending on model alignment:

\begin{itemize}
  \item \textbf{Correctly Specified Setting:} All users are recruited simultaneously to match the batch training regime assumed by \textsc{MTTS}. We simulate 50 users, each interacting for 5 rounds. We report cumulative Bayes regret after one interaction round per user (i.e., 50 global time steps).

  \item \textbf{Misspecified Setting:} Users arrive sequentially in batches of 5 over 20 recruitment rounds (100 users total). Each user interacts for 5 rounds post-arrival. This allows us to assess the adaptability of our model as it encounters more users, and to compare against the static nature of \textsc{MTTS}, which tends to overfit to early tasks.
\end{itemize}

\subsection{Compute Resources and Runtime}
All experiments were executed on a MacBook Air (2020) equipped with an Apple M1 chip and 8GB of unified memory. Code is implemented in Python using JAX and NumPy, with no GPU acceleration. Each complete experiment for each random seed runs in under 60 minutes.

\paragraph{Random Seeds.} Each experiment is repeated across fixed seeds for \( \text{seed} \in \{0, 1, \dots, 9\} \). These seeds are used to initialize environment generation, user recruitment, and baseline algorithms. For particle-based inference, additional random seeds are sampled during each round of Sequential Monte Carlo (SMC) updates to drive variability in resampling and rejuvenation steps.

\subsection{Hyperparameters (All Models)}
\begin{itemize}
  \item Number of particles: \( N = 200 \)
  \item KL expansion truncation: \( M = 80 \)
  \item Grid size: constructed via discretization
  \begin{itemize}
    \item For each arm mean \( \mu_a \in [-2, 2] \), 20 uniformly spaced grid points
    \item For context \( x \in [-1, 1] \), 10 uniformly spaced points
  \end{itemize}
  \item GP kernel: squared exponential with lengthscale = 0.7 (tuned)
  \item GP mean function: constant zero
  \item Observation noise: \( \sigma = 0.1 \)
  \item Langevin MCMC step size: \( \eta = 0.05 \)
  \item Resampling schedule: triggered every 5 SMC steps or if \( \text{ESS} < N/2 \)
  \item SMC update frequency: 5 rounds per newly observed data point
  \item Annealing: adaptive inverse temperature schedule \( \beta_t \in [0, 1] \) selected to maintain target effective sample size
\end{itemize}

\paragraph{KL Basis.} The Karhunen–Loève basis vectors used in the truncated GP expansion are computed once at initialization and fixed throughout all experiments. No per-task or per-user recomputation is performed.

\paragraph{Baselines.} All parametric baselines use closed-form posterior updates where applicable.

\end{document}